# Growing a Tail: Increasing Output Diversity in Large Language Models

Michal Shur-Ofry[1], Bar-Horowitz-Amsalem[2], Adir Rahamim[3], Yonatan Belinkov[ϒ4]

[1]Law Faculty, Hebrew University of Jerusalem; Jerusalem, Israel. michalshur@mail.huji.ac.il
[2]Law Faculty, Hebrew University of Jerusalem; Jerusalem, Israel. bar.horovitzamsalem@mail.huji.ac.il
[3]Faculty of Computer Science, Technion – Israel Institute of Technology; Haifa, Israel. adir.rahamim@campus.technion.ac.il
[4]Faculty of Computer Science, Technion – Israel Institute of Technology; Haifa, Israel. belinkov@technion.ac.il

## Abstract

How diverse are the outputs of large language models when diversity is desired? We examine the diversity of responses of several language models to questions with multiple possible answers, comparing them with human responses. Our findings suggest that models' responses are highly concentrated, reflecting narrow, mainstream outputs, in comparison to humans, whose responses exhibit a much longer-tail. We then examine three simple and practical ways to increase output diversity: 1) increasing generation randomness via temperature sampling; 2) prompting models to answer from diverse perspectives using a single prompt; and 3) aggregating outputs from several models. We find that these interventions, especially when combined, can substantially increase output diversity, although single-model outputs generally remain less diverse than the human baseline. We discuss potential implications of these findings for future work in AI policy and governance that wishes to preserve cultural diversity, an essential building block of a democratic social fabric.

[ϒ] Correspondence to Yonatan Belinkov belinkov@technion.ac.il,

## 1. Introduction

A large body of literature recognized that the mere exposure to diverse cultural and historical contents is a constructive social factor, crucial for promoting tolerance and democratic values. Conversely, a lack of diversity can result in extremism and exclusion (e.g., Benhabib, 2002; Post, 2000). Despite the importance of diversity, research in recent decades identified a trend toward uniformity and standardization in cultural markets, attributed in part to technological progress, mass media, and globalization (e.g., Baker, 2002).

How will cultural diversity fare when artificial intelligence (AI) large language models (LLMs) become a prominent source from which people derive information about the world? Recent studies in computer science discussed the notion of "algorithmic monoculture", whereby identical algorithmic outputs may lead decision makers to adopt wrong or biased decisions (e.g., Bommasani et al., 2022; Kleinberg & Raghavan, 2021). Our study's focus is not on algorithmic decisions that affect users, but rather on the exposure of users to diversity of contents ("exposure diversity") as a social end in and of itself. The significance of such diversity was recently underscored in the European Union's Artificial Intelligence Act (AI Act, 2024, Preamble 27). Yet, given that the current technological paradigm underlying LLMs relies on statistical frequencies (LeCun et al., 2015), one can expect that LLMs' default outputs will display a short tail and concentrate around the popular and mainstream items.

This article makes a focused, concrete empirical contribution. It asks whether LLMs, in ordinary user-facing settings and under prompts where multiple different answers are reasonable, generate a narrow or broad set of outputs; and it tests whether three simple interventions can increase output diversity. We do not claim that our case studies can represent all contexts in which diversity is socially valuable. Rather, we use three deliberately simple case studies---pertaining to historical figures, television series, and cities---to provide an initial evaluation of output concentration and of practical diversity-enhancing interventions.

Recent research has been exploring different aspects of diversity in text generated by LLMs, addressing LLMs tendency to produce repetitive, narrow or uniform outputs. Methods to improve diversity have focused on training procedure modifications, decoding interventions, or specialized prompting (see Section 2 for a more detailed discussion).

In this research we offer a novel integrative approach for increasing LLMs' output diversity in cases where diversity is a virtue and there are multiple possible responses. Our approach is based on integrating three straightforward measures: combining outputs from different LLMs, temperature tuning, and prompting for diversity using a direct single prompt. To evaluate the impact of each of these strategies, we first compare LLMs' default outputs to outputs generated by humans, using entropy as our principal diversity measurement. Relying on this baseline evaluation, we test the impact of each of these strategies on output diversity separately. We then integrate these methods into a single framework. This structure sheds light on the relative contribution of each of the interventions to diversity gains, and highlights the significance of combining several interventions. Importantly, the interventions we study are practical and simple, require no architectural changes, and most of them can be easily implemented by users without involvement of computer scientists or AI developers.

Thus, our objective in this research is to evaluate the diversity of outputs generated by LLMs in cases where diversity is a virtue, compare it with that of human responses, and evaluate the impact of the aforesaid interventions, both separately and combined, as a potential solution to LLMs diversity deficit. We further seek to discuss the potential implications of our findings and locate them within the broader framework of AI policy, particularly in the context of preserving cultural diversity and promoting democratic values in an increasingly AI-driven information landscape.

## 2. Related Work

Current literature on promoting diversity of LLM-generated outputs concentrates on three main approaches. First, several studies focus on modifying fine-tuning objectives or interventions (e.g., Berndsen, 2024; Chung, et al., 2025; Lanchantin, et al., 2025; Li, et al., 2025; Mai & Carson-; Zhang, et al., 2024). For instance, one of those studies proposed a method that steers models toward producing more varied responses using a deviation score in the training objective (Chung, et al., 2025). Another study similarly proposed distribution-matching fine-tuning method (Zhang et al., 2024), while yet another study introduced regularization-based methods to retain the inherent diversity in pre-trained models during fine-tuning, arguing that many current pipelines can suppress lexical and semantic variety (Li et al., 2025).

A second strand of studies explores how diversity can be enhanced through sophisticated prompting (e.g., Hayati et al., 2024; Lagzian et al., 2025; Yadav et al., 2024; Yu et al., 2023). One of the studies in this vein demonstrated how using prompts that include different attributes such as writing styles, lengths, and location, can trigger LLMs to simulate diverse perspectives that can then enrich training datasets, thereby leading to models that generate more nuanced text (Yu et al., 2023). An additional study investigated human-in-the-loop strategies, such as correcting labels, to balance diversity and accuracy (Chung et al., 2023).

Thirdly, an additional growing body of work explores decoding-level strategies to encourage diversity, such as top-k, top-p, min-p, beam-search variants, constrained sampling, and other sampling interventions to encourage diversity (Franceschelli & Musolesi, 2025; Kumar et al., 2021, 2022; Minh et al., 2025; Nadeem et al., 2020; Qin et al., 2022; Wiher et al., 2022; Wong et al., 2024; Zhang et al., 2021). Our study does not benchmark all such decoding strategies. Instead, we focus on interventions that are available or intelligible in ordinary user-facing settings, while recognizing that more advanced decoding methods are an important direction for future work.

While our study is related to the prompting and decoding lines of work, unlike parallel studies, the strategy we employ to increase LLMs' output diversity is simple, yet effective, making it practical for both users and LLM providers to adopt.

## 3. Methodology and Experimental Design

To obtain a broad outlook, we used eight LLMs available during the data collection period [i]: GPT4, Gemini, Claude V1.3, Claude-instant 1.1, J2-Mid, J2-Ultra, MPT-Chat, and GPT3.5 (for details on each of these models, see Achiam et al., 2023; Gemini Team, 2023*;* Anthropic, 2023; Lieber et al., 2021; MosaicML NLP Team, 2023). The first two models were closed models, which allow user interaction only through their user interface ("the UI models"). The additional six models have accessible APIs ("the API models") that allowed us to adapt their "temperature" – a hyperparameter that changes the randomness of the responses generated by the models (LeCun et. al, 2006).

Our study comprised two stages. At the first, baseline, stage, we set the initial temperature of the API models to 0.3.[ii] We asked each of the 8 models the following three questions:

Q1: Can you list 3 influential persons from the nineteenth century?
Q2: Can you list three good television series?
Q3: Can you list three cities worth visiting?

The questions we selected have multiple possible answers rather than a single correct one, and relate to different aspects of diversity (historical figures, cultural products, and geographical locations). We intentionally selected questions without an immediate political context or implications. Because our focus is on diversity as such, rather than on bias or algorithmic decision-making, our questions also do not have an expected "correct" distribution of answers. We assume that when people present LLMs with such seemingly mundane questions, they will likely rely on their outputs without reservations (Shur-Ofry, 2025).

We reiterated each question 10 times per model, so that the overall number of "votes" (e.g., 19th century figures, tv series, or cities) comprising the outputs was 30 votes per model.[iii] We then examined how many *different* items appear in this output. We manually reviewed outputs, normalized obvious duplicate labels referring to the same entity, and counted the number of distinct items. If a model supplied an answer outside the requested category - for example a film rather than a television series - the item was recorded and then counted as an error in the error analysis. We did not retry or replace such responses, because doing so would have introduced an additional researcher-controlled filter. We reiterated this process for each of the treatments described below. Altogether our dataset comprised approximately 3,780 "votes", which we manually reviewed and analyzed.

**Measuring output diversity:** There is no single accepted metric for cultural diversity (e.g., Napoli, 2011). We therefore used *entropy***,** which measures the level of information or surprise in possible outcomes of a random variable (Shannon, 1948), provides a good indication of how distributed (diverse) the models' responses are, and was used in other diversity studies (e.g. Lagzian et al., 2025; Li et al., 2025; Zhang et al., 2024) [iv] We also report the number of distinct responses, and variance across models for average results, with standard deviations shown in parentheses.[v]

**Internal diversity measures:** In addition, we used several internal, content-related measures, specifically tailored for each question. In coding responses to Q1, "Western" is defined as North America, Western Europe, Australia, and New Zealand; "female" is coded according to the public identity of the figure in common reference sources; and occupation categories distinguish politics/leadership, science/technology, and culture/humanities according to the figure's primary historically recognized role in common reference sources. For Q2, non-English/non-Anglophone series are series not primarily produced in the United States, Canada, or the United Kingdom; countries of origin and production year were coded according to publicly available information. For Q3, countries were coded by the country in which the city is located. Ambiguous or disputable cases were resolved conservatively and are listed in the Supplementary Materials.

**Comparison to humans:** To obtain an illustrative human baseline, we used X posts about the same questions, where people were asked to respond without using any LLMs or search engines. We considered the first 10 replies received for each question, which similarly comprised 3 votes per response. After aggregating 30 human votes from 10 responders, we similarly counted the different items included in the aggregation of votes.[vi]

In the second stage, we employed several measures to increase output diversity. First, we raised the temperature of the six API models from 0.3 to 1. Second, we adjusted the prompts by adding the words: "answer from diverse perspectives", before the original prompt in one treatment (e.g., "Answer from diverse perspectives: Can you list three good television series?") (the "diversity ante" prompt), and after the original prompt in a second treatment (e.g., "Can you list three good television series? Answer from diverse perspectives") (the "diversity post" prompt). In the API models we further combined the high temperature with the diversity-inducing prompting. Third, we aggregated the outputs of the 6 API models and of the 2 UI models under each of the diversity-inducing conditions, and applied our diversity metrics to the aggregated outputs.

Research indicates that LLMs possess the capability to simulate different worldviews, or "personas" (Andreas, 2022; Joshi et al., 2023). For instance, prompting an LLM to answer as a professor may increase its truthfulness (Lin et al., 2022), and prompting it to consider the perspective of a certain country makes responses similar to that country's population (Durmus et al., 2024). We therefore hypothesize that explicit simple prompting may increase output diversity. Furthermore, as different LLMs are trained on somewhat different datasets, they may provide different default outputs (cf., Tekin et al., 2024), or assume different default personas when prompted for diversity, and so we hypothesize that combining their responses may increase diversity.

# 4. Results

## 4.1 Low diversity of default LLM responses

In the baseline stage, all the models we examined generated highly concentrated, non-diverse, outputs, repeating a small number of popular, mainstream, items across reiterations. For example, in Q1 (19th century personae), most of the models repeatedly generated Charles Darwin, Karl Marx, and Abraham Lincoln in their responses. Out of 30 votes, and 30 potential different responses, the average number of actual different persons was 4.3 for the six API models, and 5 for the two UI models, implying that all models displayed an extremely short-tailed distribution. Conversely, the 28 human votes on 'X' to the same question comprised 25 different figures out of a possible 30, displaying a very long tail in comparison. Very similar patterns emerged with respect to the outputs generated for Q2 (television series) and Q3 (cities): Despite the breadth of potential outputs, the actual outputs generated by the AI models to each of these questions concentrated around a small number of repeating highly-popular items, displaying a very short tail, while the human responses on 'X' were significantly more diverse and distributed, as illustrated in *Figure 1*. The complete data generated by the AI models and humans in response to each of the questions appear in the Supplementary Materials, Parts II-IV.

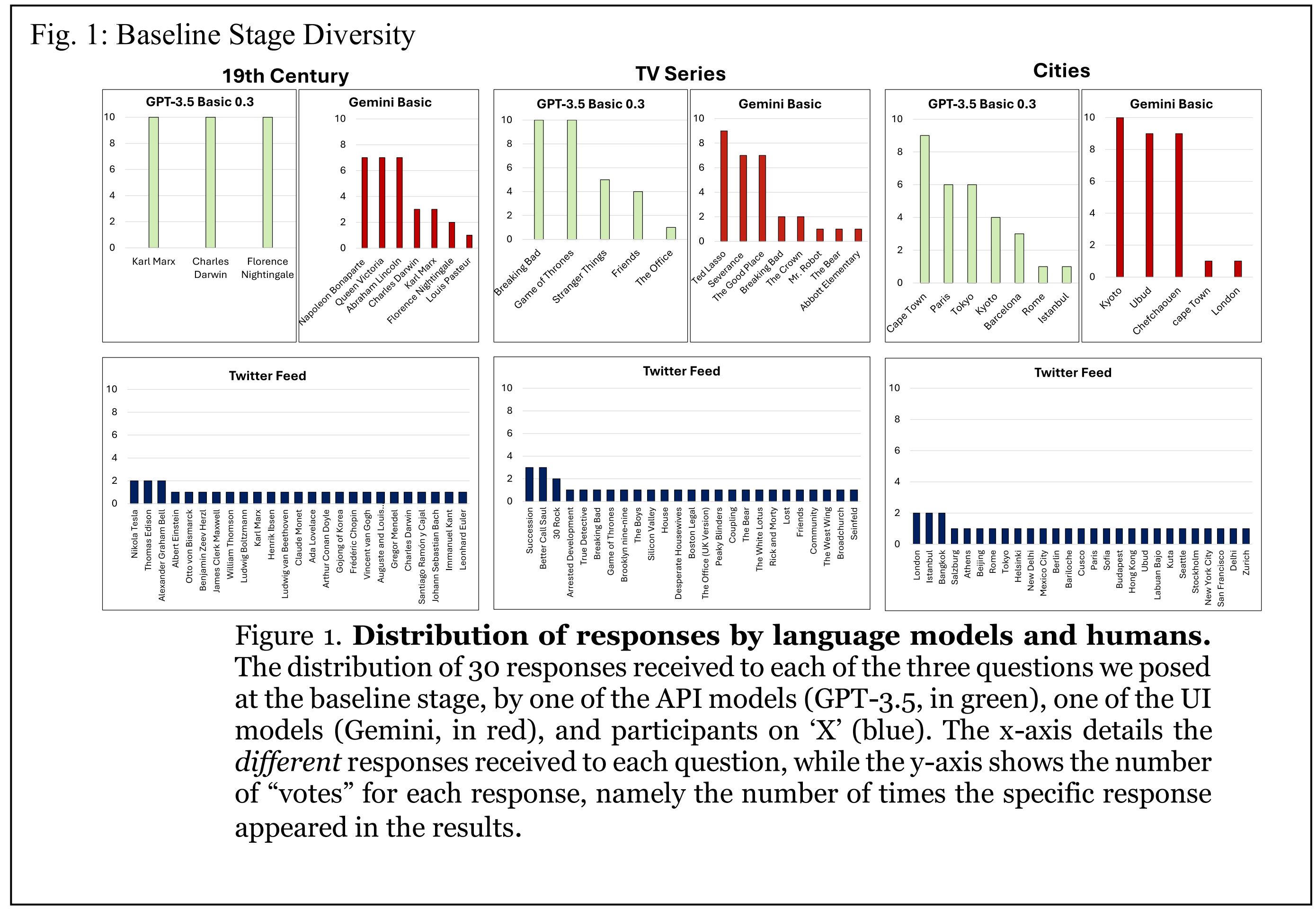

Figure 1. **Distribution of responses by language models and humans.** The distribution of 30 responses received to each of the three questions we posed at the baseline stage, by one of the API models (GPT-3.5, in green), one of the UI models (Gemini, in red), and participants on 'X' (blue). The x-axis details the *different* responses received to each question, while the y-axis shows the number of "votes" for each response, namely the number of times the specific response appeared in the results.

Accordingly, the entropy levels of the models under the basic treatment for all three questions were relatively low, with averages ranging between 2.1 and 2.9. The entropy of the human responses was significantly higher, ranging between 4.6 and 4.7, as summarized in Table 1. The differences are much larger than the reported standard deviations. The number of unique responses (Table 2) tells a similar story: much higher numbers in human responses compared to model responses.

| | Human ['X'] | | 6 API MODLES | | | | | | 2 UI MODELS | | |
|---|---|---|---|---|---|---|---|---|---|---|---|
| | | | Base | Temp 1.0 | Prompt D-ante | Prompt D-post | Prompt D-ante & Temp 1.0 | Prompt D-post & Temp 1.0 | Base | Prompt D-ante | Prompt D-post |
| 19TH CENTURY | 4.6 | AVE | 2.1 | 2.5 | 2.3 | 2.5 | 3.0 | 3.4 | 2.4 | 3.8 | 2.8 |
| | | STD | 0.3 | 0.5 | 0.4 | 0.5 | 0.6 | 0.2 | 0.4 | 0.7 | 0.6 |
| | 0.4 | AGG | 3.0 | 3.1 | 3.3 | 3.6 | 3.7 | 4.2 | 2.7 | 4.2 | 3.8 |
| TV SERIES | 4.6 | AVE | 2.6 | 3.1 | 2.3 | 2.9 | 3.4 | 4.0 | 2.4 | 3.0 | 2.8 |
| | | STD | 0.5 | 0.9 | 0.3 | 0.5 | 0.8 | 0.5 | 0.4 | 0.3 | 0.3 |
| | 0.6 | AGG | 3.2 | 3.9 | 4.0 | 3.9 | 4.8 | 5.1 | 3.1 | 3.9 | 3.8 |
| CITIES | 4.7 | AVE | 2.9 | 2.9 | 2.5 | 2.4 | 3.4 | 3.4 | 2.2 | 3.4 | 2.6 |
| | | STD | 0.4 | 0.7 | 0.3 | 0.2 | 0.4 | 0.6 | 0.1 | 0.5 | 0.7 |
| | 0.3 | AGG | 2.8 | 3.4 | 3.4 | 3.4 | 4.3 | 4.0 | 2.7 | 4.2 | 3.1 |

Table 1. **Diversity of responses.** The entropy levels of the responses attained by the API models (in green), the UI models (red) and participants on 'X' (blue), under each of the different conditions, for each of the three questions. The columns represent the different treatments. The "AVE" lines specify the average entropy of the responses generated by each group of models (with standard deviations below), under each treatment. The "AGG" line specifies the entropy of responses generated when aggregating each group of models under each treatment. The colors are conditioned in a gradient way, so that sharper colors reflect higher entropy.

**Human response extension check:** While we limited the human responses to 10 respondents to preserve comparability with the 10 iterations used for each model, the general lack of overlap among the human replies implies that increasing the human sample could likely expand the human-response tail even further. We therefore added a supplemental check, and inspected additional responses received in the original X threads. As detailed in the Supplementary Materials, in Q1, three additional responses comprised nine distinct figures, seven of which were not in the original dataset. In Q2, three additional responses comprised nine distinct series, six of which were not in the original dataset. In Q3, four additional responses comprised twelve city mentions, eight of which were not in the original dataset. This suggests that increasing the human sample would likely increase the human-response tail and widen the gap between the humans' and the models' entropy even further.

### 4.2 Diversity-promoting measures

How did output diversity of LLMs change when employing diversity-increasing measures? Tables 1 and 2 summarize entropy and number of responses from employing the different diversity-promoting measures. More detailed results appear in the Supplementary Materials, Parts II-IV.

| | | | Base | Temp 1.0 | Prompt D-ante | Prompt D-post | Prompt D-ante & Temp 1.0 | Prompt D-post & Temp 1.0 | Base | Prompt D-ante | Prompt D-post |
|---|---|---|---|---|---|---|---|---|---|---|---|
| 19$^{TH}$ CENTURY | 25 /28 | AVE | 4.3 /30 | 7.8 /29 | 5.3 /30 | 6.7 /30 | 11.3 /29 | 13.3 /30 | 5.0 /30 | 9.0 /30 | 7.5 /30 |
| | | STD | 1.1 | 2.3 | 2.5 | 4.4 | 2.3 | 1.5 | 0.0 | 1.0 | 3.5 |
| | | AGG | 10.0 /180 | 19.0 /171 | 16.0 /180 | 21.0 /180 | 33.0 /176 | 37.0 /180 | 8.0 /60 | 16.0 /60 | 13.0 /60 |
| TV SERIES | 25 /30 | AVE | 5.7 /30 | 11.0 /30 | 5.8 /30 | 9.2 /30 | 14.2 /30 | 17.5 /30 | 6.5 /30 | 9.5 /30 | 10.0 /30 |
| | | STD | 2.1 | 7.7 | 1.5 | 5.7 | 2.9 | 4.8 | 1.5 | 2.0 | 1.5 |
| | | AGG | 17.0 /180 | 37.0 /180 | 22.0 /180 | 33.0 /179 | 58.0 /180 | 66.0 /180 | 11.0 /60 | 18.0 /60 | 20.0 /60 |
| CITIES | 27 /30 | AVE | 5.2 /30 | 9.8 /29 | 7.0 /30 | 7.0 /30 | 13.5 /30 | 13.6 /30 | 5.5 /30 | 8.5 /30 | 7.5 /30 |
| | | STD | 1.2 | 5.3 | 2.4 | 2.2 | 1.8 | 5.0 | 0.5 | 2.5 | 0.5 |
| | | AGG | 14.0 /180 | 30.0 /176 | 19.0 /180 | 21.0 /180 | 44.0 /177 | 41.0 /179 | 9.0 /60 | 22.0 /60 | 13.0 /60 |

Table 2. **Number of responses.** The aggregate number of responses attained by the API models (in green), the UI models (red) and participants on 'X' (blue), under each of the different conditions, for each of the three questions. The columns represent the different treatments. The "AVE" lines specify the average numbers of *different* responses generated by each group of models (with standard deviations below), under each treatment. The smaller number appearing in the lower part of the square is the average number of "votes" generated by that group of models under the specific treatment. The "AGG" line specifies the numbers of *different* responses generated when aggregating the entire group of models, while the smaller numbers appearing in the lower part of the square are the total number of "votes" generated by the aggregation of that group of models. The colors are conditioned in a gradient way, so that sharper colors reflect greater diversity of responses.

As is evident from Table 2, raising the temperature of the API models from 0.3 to 1 increased the variety of responses by roughly 80%-100%. For example, in Q1 ($19^{th}$ century figures), the average number of different responses increased from 4.3 to 7.8 (out of a possible 30). Increases in diversity under maximum temperature further occurred in Q2 and Q3, where, similarly, the number of different items roughly doubled.

Prompting the models to answer from diverse perspectives increased the variety of the responses by approximately 20% to 60%, relative to the baseline stage. For example, in Q1 ($19^{th}$ century figures), the average number of different persons in the API models increased from 4.3 under the baseline condition, to 5.33 under the diversity-ante prompt, and 6.66 under the diversity-post prompt. The average number of different persons in the UI models' outputs similarly increased from 5 under the baseline condition, to 9 under diversity-ante prompt, and 7.5 under the diversity-post prompt. Roughly similar increases in diversity under the two diversity-inducing prompts occurred in Q2 and Q3.[vii]

Combining the diversity prompting with a maximum temperature in the API models proved more effective than each of these measures alone, resulting in an increase of roughly 150%-200% in the variety of the responses, relative to the baseline.

**Internal diversity measures:** Under the diversity-inducing treatments, we have also seen limited increases in some of the internal, content-related diversity measures, which we tailored to each question. To illustrate, in Q1 ($19^{th}$ century figures), the average percentage of female figures in the outputs of the API models increased from 17% under the baseline condition, to 21% under the maximum-temperature condition, 25% under the diversity-post prompting, and 30% under the combined treatment of maximum-temperature plus diversity-post prompting. The percentage of non-western persons in the UI models' outputs increased from zero under the baseline condition, to 2-3% under the diversity-increasing prompting. We note, however, that increases in internal diversity measures were often slight, and did not occur with respect to all questions and parameters. For example, in Q2 (tv series) most models did not generate a greater number of older series under the diversity-increasing treatments. Additionally, even under the diversity-inducing treatments, the responses generated by the models, often reflected well-known, popular, and mainstream items (particularly in Q1 and Q2). The complete data pertaining to internal diversity measures appear in the Supplementary Materials, Part I-4.

Despite the measures described above, none of the models' outputs reached diversity levels comparable to those attained by humans responding on 'X': Even under combined treatment of diversity-prompting plus maximum temperature, the variety of human responses was roughly twofold greater than the models' averages (see Table 2), and the entropy of human responses remained substantially higher than that of the models (see Table 1). We therefore turned to examine whether aggregating several models can further improve output diversity, and analyzed the aggregations, rather than averages, of responses under each condition. We aggregated the 6 API models, and the 2 UI models, in two separate groups.

**Aggregation of models:** Under the baseline condition, the aggregation of models' outputs did not substantially change the diversity picture. For example, in Q1 ($19^{th}$ century figures), the 60 aggregated votes of the UI models generated only 8 different persons, a number identical to the average output of those models. Similarly, the 180 aggregated votes of the 6 API models generated only 10 different persons. This implies that the models' default outputs were overlapping to a large extent. In Q2 and Q3 the models' aggregated outputs (approximately 60 outputs of the `UI models and 180 outputs of the API models) yielded a somewhat greater variety of items, but still did not reach a level comparable to the 30 human responses on 'X' (see Table 2). However, combining the diversity inducing measures--namely high temperature, diversity-inducing prompts, or both--with an aggregation of models, often increased output diversity to a level in the range of, and in some cases exceeding, the diversity of 30 human responses on 'X'. For example, in Q2 (tv series) the aggregation of 2 UI models with a diversity-inducing prompts yielded 18 different series (under diversity-ante) or 20 different series (under diversity-post), compared to 25 different series on 'X'. The aggregation of 6 API models with the diversity-inducing treatments plus a high temperature resulted in 58 different series (under diversity-ante) or 60 different series (under diversity-post), numbers that exceed the number of human responses (see Table 2). These results imply that aggregating models while using diversity inducing treatments decreases the overlap between the outputs of different models. The aggregated outputs in such cases further displayed a long-tail distribution, as illustrated in Figure 2. The complete data pertaining to these distributions appear in the Supplementary Materials, Part I Section 5, and Parts II-IV.

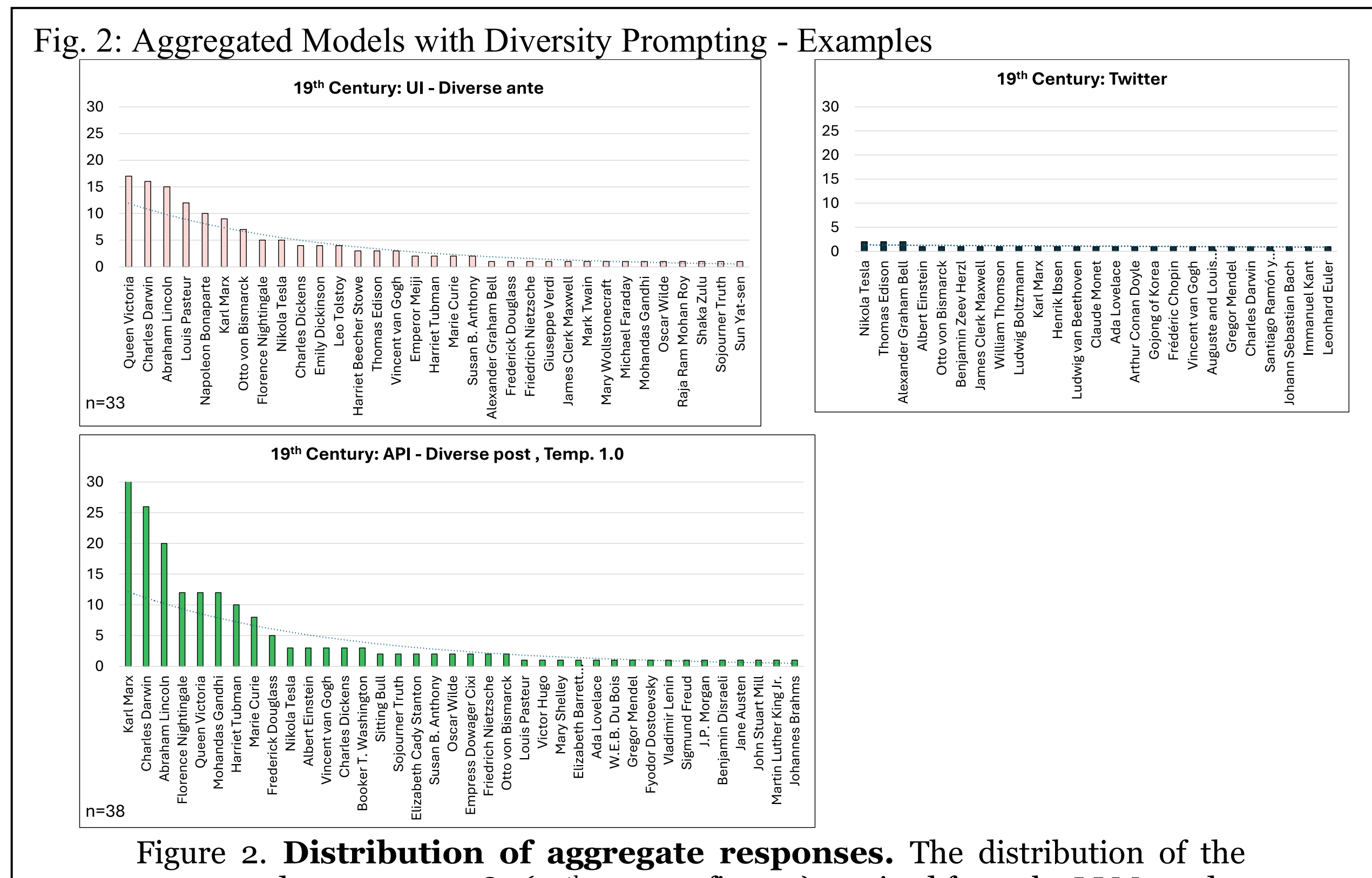

Figure 2. **Distribution of aggregate responses.** The distribution of the aggregated responses to Q1 (19th century figures) received from the LLMs under 'combined treatments' conditions, relative to the distribution of responses to that question by participants on 'X'. The upper-left image (red) depicts the distribution of approximately 60 aggregated "votes" received from 2 UI models, under a diversity-ante prompting. The bottom image (green) depicts the distribution of approximately 180 aggregated 'votes' received from 6 API models, under a diversity-ante prompting. The upper-right image (blue) depicts the distribution of 30 aggregated votes received from participants on 'X'.

Notably, even under the "aggregated" condition, the ratio between the number of different items generated by the models and the overall number of votes remains substantially low relative to the ratio in the human replies. Likewise, in the vast majority of cases, the entropy of the human responses comprising a total of 30 votes still exceeds the entropy of the aggregated models' responses, comprising a total of 180 (for API models) or 60 (for UI models) votes. Therefore, the aggregated results do not attest to the diversity of the models in-and-of themselves. Rather, they highlight that each models reflects a somewhat different range of outputs, and attest to the significance of combining several models.

Finally, we also monitored the numbers of errors generated under each condition (examples for errors are, e.g., naming William Shakespeare who lived in the 16th century in response to Q1, or naming a film rather than a television series in response to Q2). One concern with promoting for diversity is the generation of more erroneous answers, given previous findings that accuracy and diversity of language model generations can be incongruent (Nadeem et. al, 2020; Zhang et al., 2021). Interestingly, we did not find that the diversity-inducing treatments substantially increased the percentage of errors in the models' outputs. The relevant errors-data appear in the Supplementary Materials, Part I, Section 6.

## 5. Discussion

Our findings yield several primary insights. First, our case studies indicate that the default outputs of LLMs, in cases where diversity is warranted, reflect a narrow and extremely concentrated range of outputs, rather than a multiplicity of options. The models' outputs in all our baseline treatments displayed a very short tail, converging around a small number of popular and well-known items. Due to significant overlaps among the outputs of different models, aggregating default outputs from several models was insufficient to substantially increase output diversity. Human responses displayed a much broader distribution and higher levels of entropy in comparison to all the models we examined. This result is not entirely surprising, given that the main technological paradigm underlying LLMs relies on statistical probability: the frequency of an item's occurrences in the underlying training datasets crucially influences the generated outputs. Therefore, even when a multiplicity of responses is desirable, as is often the case in cultural, historical, and geographical contexts, LLMs outputs are likely to be geared toward the mean, standard, and popular.

Second, despite the concentrated outputs in the models' default states, a combination of fairly simple measures, namely temperature increase and diversity-inducing prompts, can substantially increase output diversity. Interestingly, combining the two diversity inducing measures (temperature plus prompting) produced "a whole that is greater than the sum of its parts", namely the difference between the combined measures and the default state is larger than the sum of the differences between each individual measure and the default state.

Third, even after combining maximum temperature with diversity prompting, none of the models alone generated a long tail of outputs that was comparable to, or exceeded, the long tail of outputs generated by humans. Achieving higher diversity levels required aggregating the outputs of several models, while applying one or more diversity-inducing measures. In other words, although all the models we examined displayed a rather narrow output distribution, their aggregation, while applying other diversity-inducing measures, projected a broader "slice" of the world. This latter insight highlights the importance of access to different language models, rather than a single one.

Our inquiry has several limitations. First, while the three case studies we selected pertain to different aspects of cultural diversity, they are limited and of course not exhaustive. Extending our findings to additional contexts, and to additional AI models that are not language-based (e.g., image or video generators) requires further research.

Second, our use of 'X' threads for evaluating diversity among humans is limited in several respects. The human respondents in the threads are limited in number and do not constitute a representative sample of human society. However, our main focus in this study was not the contents of the responses but rather on their diversity and distribution--e.g., we were not exploring which particular figures people deem as the most important in the 19th century, but rather how distributed (or concentrated) their answers are. We further note that previous works analyzing people's cultural choices, including comparable threads on social media accounts, similarly found long tail distributions (Anderson, 2006; Shur-Ofry, 2025). We also acknowledge that attaining responses from 10 different people on 'X' is not entirely equal to obtaining 10 reiterations from the same language model. Yet, while a perfect comparison between humans and LLMs is likely impossible, our method of 10 reiterations per model roughly simulates real-world scenarios, whereby different users who approach a language model with similar questions will receive largely uniform responses, despite the broad variety of potential answers. Therefore, the aggregation of models' responses is indicative of a potential decline in diversity.

In addition, our data were collected between July 2023 and February 2024. Because some UI models and closed models may change over time in ways that are not fully transparent, exact replication may be difficult. This is a general challenge in research on deployed LLM systems. We therefore provide the data-collection dates, model-access details available to us, prompts, and raw output materials in the supplementary files to improve reproducibility. We propose that our study be viewed as a snapshot of that motivates further continuous work rather than a comprehensive benchmark. We also highlight periodic inquiries into diversity levels of new LLM versions as a future desideratum for both research and AI policy.

Lastly, while our diversity-inducing measures were undoubtedly efficient in producing longer tails of outputs, they did not consistently generate a significant improvement in content-related parameters. For example, in Q2, the television series outputs under all treatments comprised almost exclusively English-speaking series. This implies that improving the content-related aspects of output diversity, in a way that would expose users to both local and foreign outputs, contemporary and past outputs, high and popular culture, etc., is a more complicated challenge (Baker, 2002; Benhabib, 2002; Corse, 1997; Napoli, 2011; Post, 2000). Relatedly, we note that the diversity-inducing measures we applied are intentionally simple. More technologically advanced methods, such as chain-of-thought prompting and its derivatives (Wei at al., 2022), different decoding techniques (Franceschelli et al., 2025; Kumar et al., 2021; Kumar et al., 2022; Minh et al., 2025; Qin et al., 2022; Wiher et al., 2022; Wong et al., 2024), or access to external tools and knowledge bases (Qu et al., 2024; Schick et al., 2023) may prove more effective in improving levels of diversity along additional dimensions. We mark this as a topic for future research.

## 6. Policy Implications and Conclusion

The study suggests policy-relevant implications, but these implications should be read cautiously. The experiments are limited in scope and do not by themselves establish systemic effects on democracy, social tolerance, or cultural markets. They do, however, identify a mechanism worthy of governance attention: in ordinary open-ended queries, LLMs may expose users to a narrower set of answers than the range available in human responses, unless diversity-enhancing design choices are used.

Our results indicate that, despite being trained on vast amounts of materials, LLMs are not geared toward diversity in their default interactions with users, but rather toward standardization, conformity and mainstream. Since the outputs of language models are likely to influence their users' perceptions (Shur-Ofry, 2025), in the long term this raises concerns of a general decline in cultural diversity.

Yet, our study further implies that addressing the challenge of maintaining diversity does not necessitate extremely complicated or expensive interventions. Our results show that diverse contents are embedded within the models. A few simple and cheap measures, such as temperature increase and diversity-inducing prompting, can 'extract' these contents and significantly improve diversity levels. AI developers can easily incorporate such elements as design features, available to users in a transparent and accessible way. Policymakers could adopt measures to encourage developers to take such steps, for example, by explicitly including diversity as a high-level principle in AI ethical guidelines and regulation (similar to other acknowledged principles such as 'transparency', 'data security', etc.). Concomitantly, users who are aware of language models' propensity toward uniformity can use those simple measures to induce more diverse outputs. Our findings therefore highlight the importance of promoting AI literacy among general audiences, to encourage such proactive and informed uses.

Finally, our findings indicate that the aggregation of several LLMs substantially increases output diversity. Even if an individual user is unlikely to consult with numerous models, the availability of several models in the market would expose different users to different outputs, and could increase societal diversity in the aggregate. Therefore, our study provides some grounds for the need to ensure competition and diversity in the market for language models.

More generally, we frame our paper as an empirical inquiry and a call for broader research, rather than as a complete policy prescription. Subject to this cautionary note, our research marks cultural diversity as a societal goal that should be explicitly addressed by stakeholders and policymakers in the field of AI, in order to prevent systemic harm to social tolerance, solidarity, and democracy.

[i] We conducted the study over eight months between July 2023 and February 2024, and used representative LLMs available during that period. For a detailed timetable of data collection, see Supplementary Materials, Part I, Section 2. The UI models are the versions available through the user interface at the time of data collection, and the precise UI-side temperature and system settings are not publicly known. Some of the models may have undergone certain developments during the study's period, (e.g., changes in training materials), unbeknownst to us. Such possible changes, if occurred, may have affected the uniformity of outputs in specific cases, but are unlikely to have had a substantial effect on our inquiry, which focuses on general patterns, and not on specific outputs.

[ii] The default user interface temperature in these models is not officially published.

[iii] Some of the models' replies included fewer than three votes, despite our prompting. In such cases the aggregate number of votes was slightly lower than 30, as detailed in Table 2 and the Supplementary Materials, Parts II-IV.

[iv] For details and explanations on the measurement of entropy, see Supplementary Materials, Part I, Section 1-b.

[v] We considered applying complementary lexical and semantic metrics such as Self-BLEU and sentence-embedding similarity, but these metrics are designed primarily for comparing generated text strings. Our output units are normalized entity names rather than free-form

text passages, so lexical similarity would often measure spelling or alias choices rather than cultural diversity.

[vi] Links to the relevant X's threads are in the Supplementary Materials, Part I, Section 3. In Q1, one of the responses on 'X' included 1 rather than 3 votes, so that the aggregate number of actual human votes for that question was 28.

[vii] Interestingly, in the UI models group the increase in diversity was greater under the "diversity post" prompt relative to the "diversity ante" prompt, whereas in the API models group the increase was greater when the prompting inducing diversity preceded the actual question ("diversity ante").

# Supplementary Materials for

# Growing a Tail: Increasing Output Diversity in Large Language Models

# Part I

## Design Details

### *Questions*

The three questions we used in the study, defined as Q1, Q2 and Q3, were the following:

Q1: Can you list 3 influential persons from the nineteenth century?
Q2: Can you list three good television series?
Q3: Can you list three cities worth visiting?

### *Entropy*

Our entropy calculation aims to assess the diversity in responses from a given model to a given question, with a certain combination of prompting (base, post, or ante) and temperature. The larger the number of different answers, the greater should be the diversity. However, a naïve calculation of entropy by only considering the set of answers for a given combination of model-prompt-temperature may not allow for fair comparison, since the set of answers for each such combination may be different. Instead, we wish to consider the set of all possible answers produced by any combination of model, prompt, and temperature for a given question. To allow for unseen events in the entropy calculation (e.g., answers that were not produced by a given model by were given by a different model), we "smooth out" the answer counts by adding a small number. Concretely, we assign such cases a value of 1/(num_answers), where "num_answers" is the number of all possible answers produced by any model, prompt, and temperature. In practice, this modification primarily affects low-diversity settings by slightly increasing their entropy, while high-diversity settings remain largely unaffected.

**Data Collection Dates**

The table below includes the dates on which we generated output from the various models, categorized by the different research questions and, in some instances, the different scenarios.

| Date | LLM's Model & Question |
|---|---|
| Monday, July 10, 2023 | J2-mid, J2-ultra, Claude V1.3 & V1.1 Q1 |
| Wednesday, August 9, 2023 | GPT-3.5 Q1 |
| Wednesday, July 19, 2023 | MPT Q1 |
| Monday, February 5, 2024 | 6 API models Q3 |
| Monday, February 5, 2024 | 6 API models Q2 |
| Sunday, July 30, 2023 | GPT-4 Q1 - Basic, Diverse post |
| Tuesday, January 30, 2024 | GPT-4 Q3 |
| Monday, February 5, 2024 | 6 API models Q1 |
| Tuesday, February 20, 2024 | 6 API models Q2 |
| Tuesday, February 20, 2024 | 6 API models Q3 |
| Thursday, February 22, 2024 | GPT-4 Q1 - Diverse Ante |
| Thursday, February 22, 2024 | Gemini Q1 |
| Thursday, February 22, 2024 | Gemini Q3 - Diverse Ante |
| Sunday, February 25, 2024 | Gemini Q3 - Basic, Diverse Post |
| Saturday, February 24, 2024 | Gemini Q2 |
| Saturday, February 24, 2024 | GPT-4 Q2 |

**X's threads**

Below are the links to the original X (formerly Twitter) threads that contain the questions addressed in the study, and the responses thereto.

Q1 – https://x.com/boknilev/status/1679470574270050312?s=20.
Q2 – https://x.com/boknilev/status/1759613863111061902.
Q3 – https://x.com/boknilev/status/1757406317835145661.

**Human-response extension check**: We inspected later X-thread replies beyond the first 10 respondents.

Q1 added three responses comprising the following nine distinct figures, out of which seven were not included in the original dataset: Marie & Pierre Curie (counted as a single reply), Swami Vivekananda, Abe Lincoln, Albert Einstein, Alexander Fleming, Napoleon Bonaparte; Benjamin D'Israeli, Victor Hugo, Bismarck.

Q2 added three responses comprising the following nine distinct series, out of which following six were not included in the original dataset: Vikings, Peaky Blinders, the Black List, Brooklyn 99, the Sopranos, the Americans, Breaking Bad, Severance, The Simpsons.

Q3 added four responses comprising the following twelve city mentions, out of which eight were not included in the original dataset: Palo Alto, New York City, Austin, Lisbon, Osaka, Florence, Malaga, Athens, Rome, Dublin, New York, Hyderabad.

This extension implies that increasing the human sample would likely widen rather than narrow the diversity gap between the human baseline and LLM outputs.

### Internal Diversity Measures

The charts include variables defined as Diversity Measures, encompassing various non-numeric characteristics relevant to each type of research question, including demographic attributes, geographic information, cultural aspects etc.

### *Q1*

| | API Models | | | | | | UI Models | | |
|---|---|---|---|---|---|---|---|---|---|
| | Basic (with 0.3 Temperature) | Basic (with 1.0 Temperature) | Diverse post (with 0.3 Temperature) | Diverse post (with 1.0 Temperature) | Diverse Ante (with 0.3 Temperature) | Diverse Ante (with 1.0 Temperature) | Basic | Diverse Post | Diverse Ante |
| Avg. non-western[1] | **9%** | **5%** | **7%** | **12%** | **11%** | **11%** | **0%** | **2%** | **3%** |
| Avg. Females | **17%** | **21%** | **25%** | **30%** | **21%** | **22%** | **32%** | **33%** | **37%** |
| Avg. Politics & Leadership[2] | **34%** | **41%** | **46%** | **51%** | **46%** | **43%** | **32%** | **35%** | **38%** |
| Avg. Science & Technology[3] | **42%** | **36%** | **28%** | **25%** | **27%** | **31%** | **50%** | **33%** | **45%** |
| Avg. Culture & Humanities[4] | **24%** | **23%** | **26%** | **24%** | **27%** | **27%** | **18%** | **18%** | **13%** |

[1] We defined "Western" as including North America, Western Europe, Australia, and New Zealand.

[2] "Politics & Leadership" refers to individuals whose primary occupations were in politics and various leadership positions. This includes politicians, government officials, and leaders in both public and private sectors.

[3] "Science & Technology" refers to individuals whose primary occupations were in the fields of scientific research, technological development, and related disciplines. This includes scientists, engineers, researchers, and technologists.

[4] "Culture & Humanities" refers to individuals whose primary occupations were in cultural and humanistic disciplines. This includes artists, writers, philosophers, historians, and other professionals engaged in the exploration and expression of human culture, values, and historical contexts. In coding the responses in all of Internal Diversity Measures we used common references sources.

## *Q2*

| | API Models | | | | | | UI Models | | |
|---|---|---|---|---|---|---|---|---|---|
| | Basic (with 0.3 Temperature) | Basic (with 1.0 Temperature) | Diverse post (with 0.3 Temperature) | Diverse post (with 1.0 Temperature) | Diverse Ante (with 0.3 Temperature) | Diverse Ante (with 1.0 Temperature) | Basic | Diverse Post | Diverse Ante |
| Earliest Production year | **1994** | **1989** | **1959** | **1959** | **1994** | **1989** | **2005** | **1999** | **1959** |
| Avg. non-English[5] | **0%** | **1%** | **0%** | **3%** | **1%** | **1%** | **0%** | **0%** | **2%** |
| Avg. No. of countries of origin | **1** | **2** | **1.83** | **2.83** | **1.83** | **2** | **1.5** | **2** | **3** |
| Avg. Old Series[6] | **11%** | **10%** | **10%** | **12%** | **11%** | **10%** | **0%** | **7%** | **10%** |

## *Q3*

| | API Models | | | | | | UI Models | | |
|---|---|---|---|---|---|---|---|---|---|
| | Basic (with 0.3 Temperature) | Basic (with 1.0 Temperature) | Diverse post (with 0.3 Temperature) | Diverse post (with 1.0 Temperature) | Diverse Ante (with 0.3 Temperature) | Diverse Ante (with 1.0 Temperature) | Basic | Diverse Post | Diverse Ante |
| Avg. No. of countries | **5** | **8** | **7** | **11** | **7** | **10** | **5** | **7** | **9** |

[5] "Non-English" refers to TV series that are not American, Canadian or British productions.

[6] "Old series" are defined as television series or programs that commenced production prior to the year 2000.

**Aggregated Models' Distributions** –
The figures below depict aggregated data from six API models, two UI models, and Twitter, under each of the defined treatments.

## *Q1:*

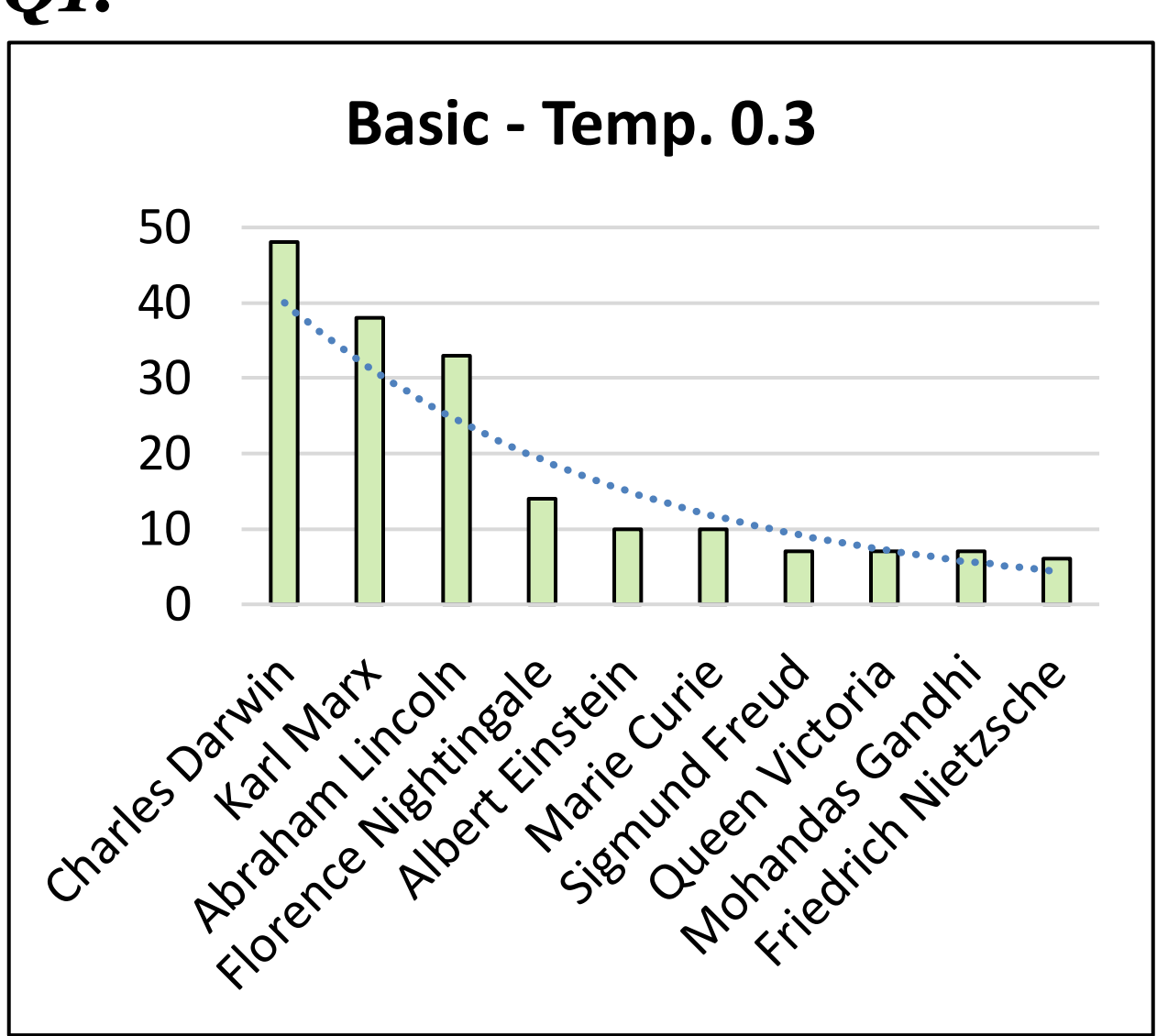

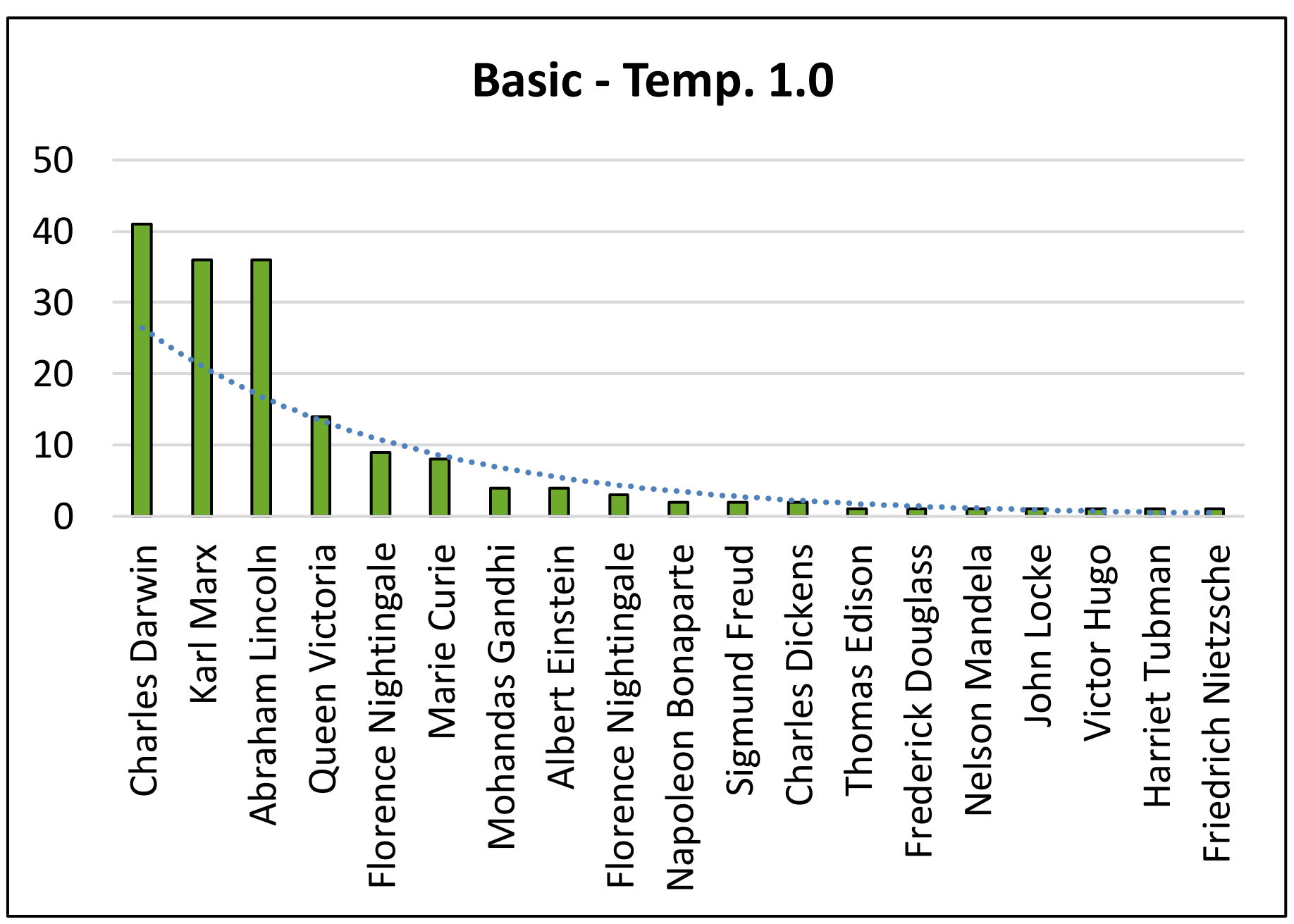

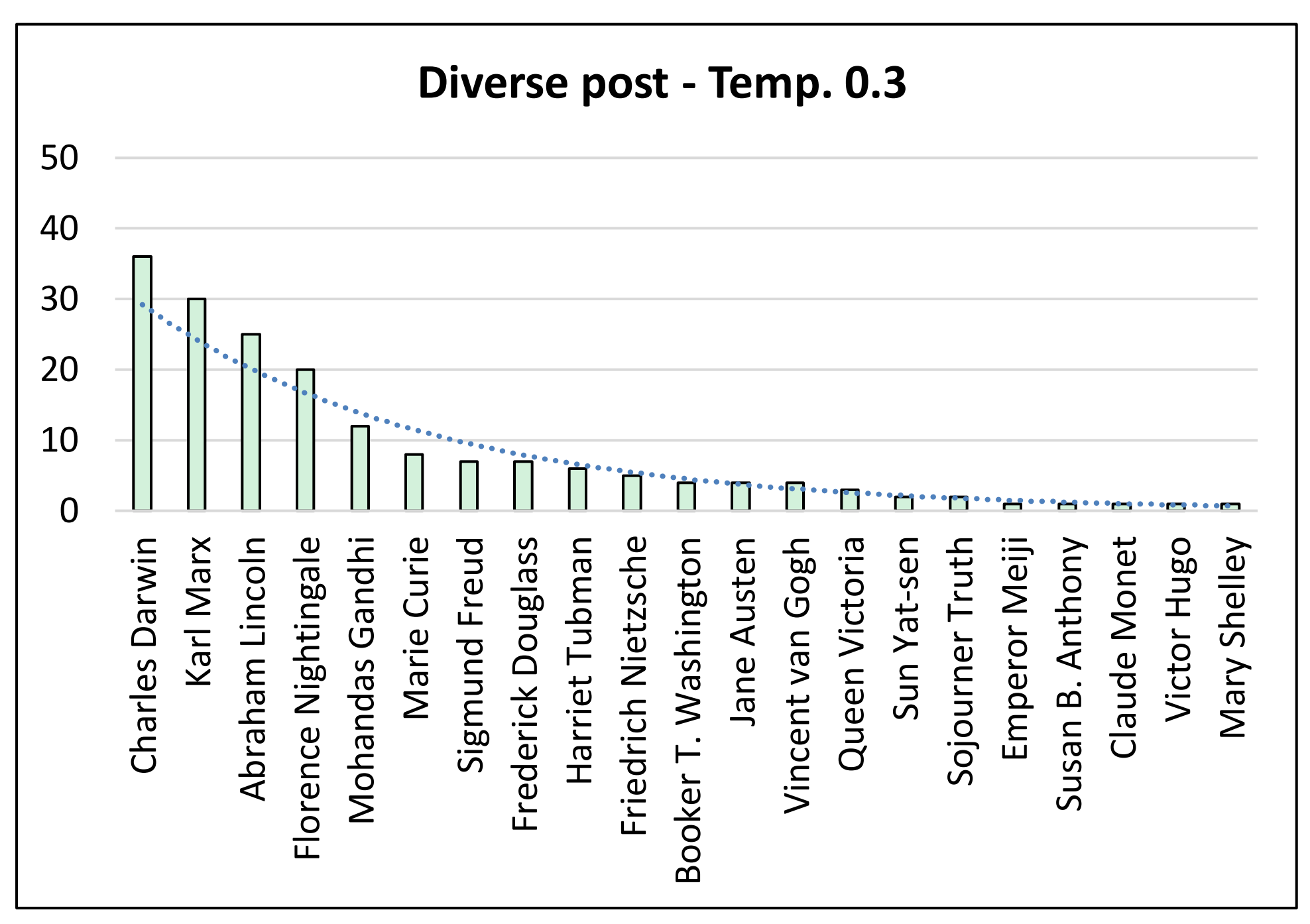

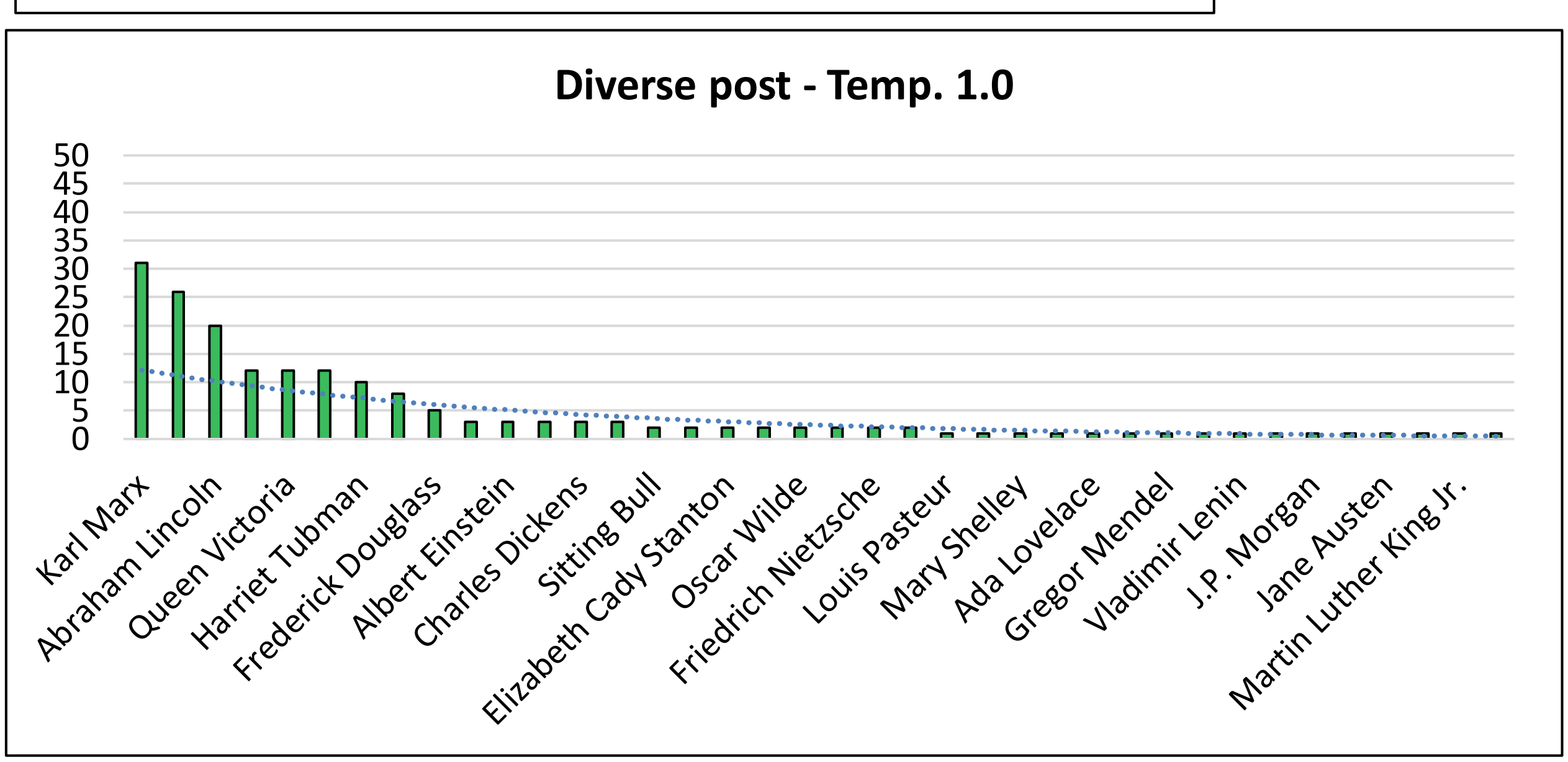

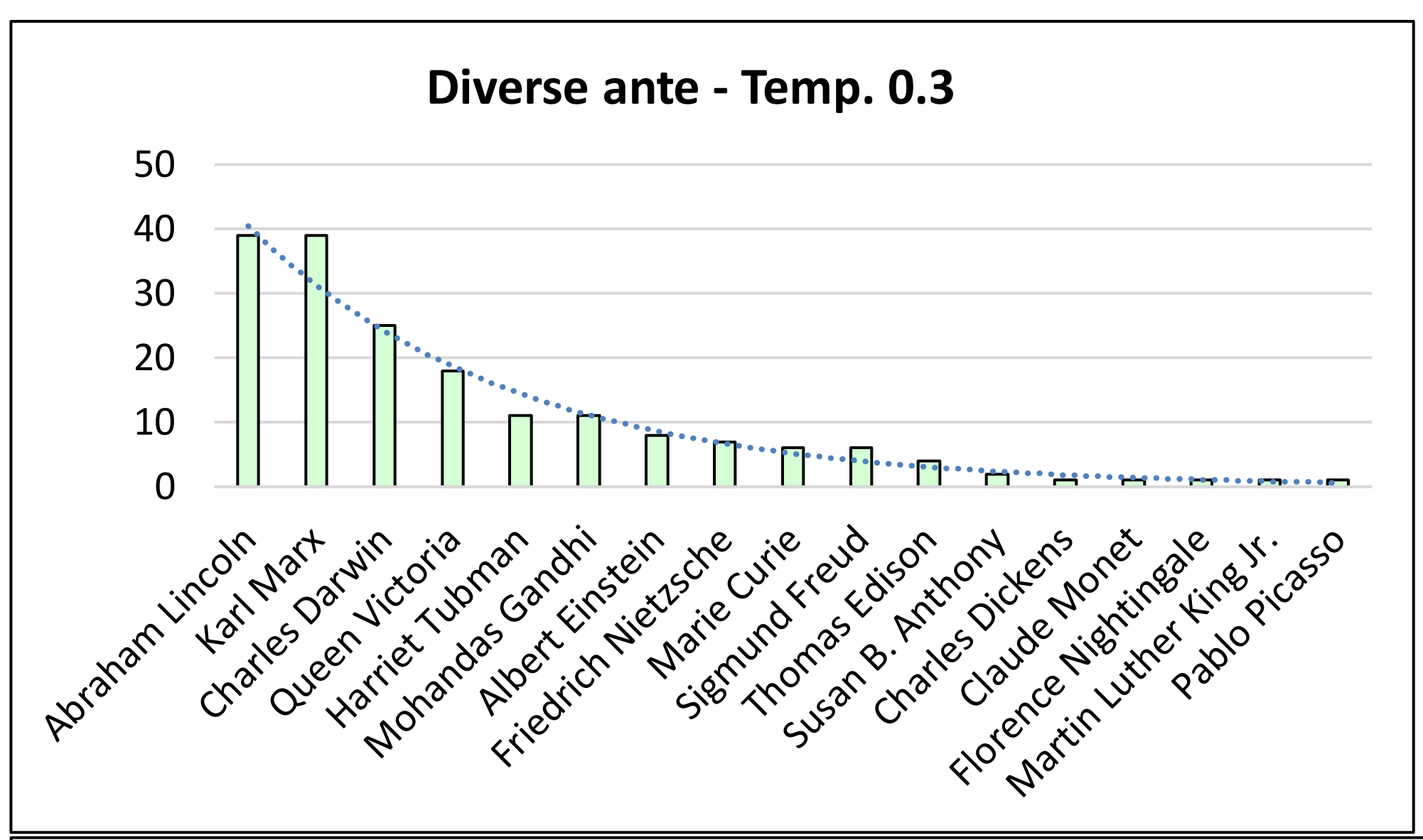

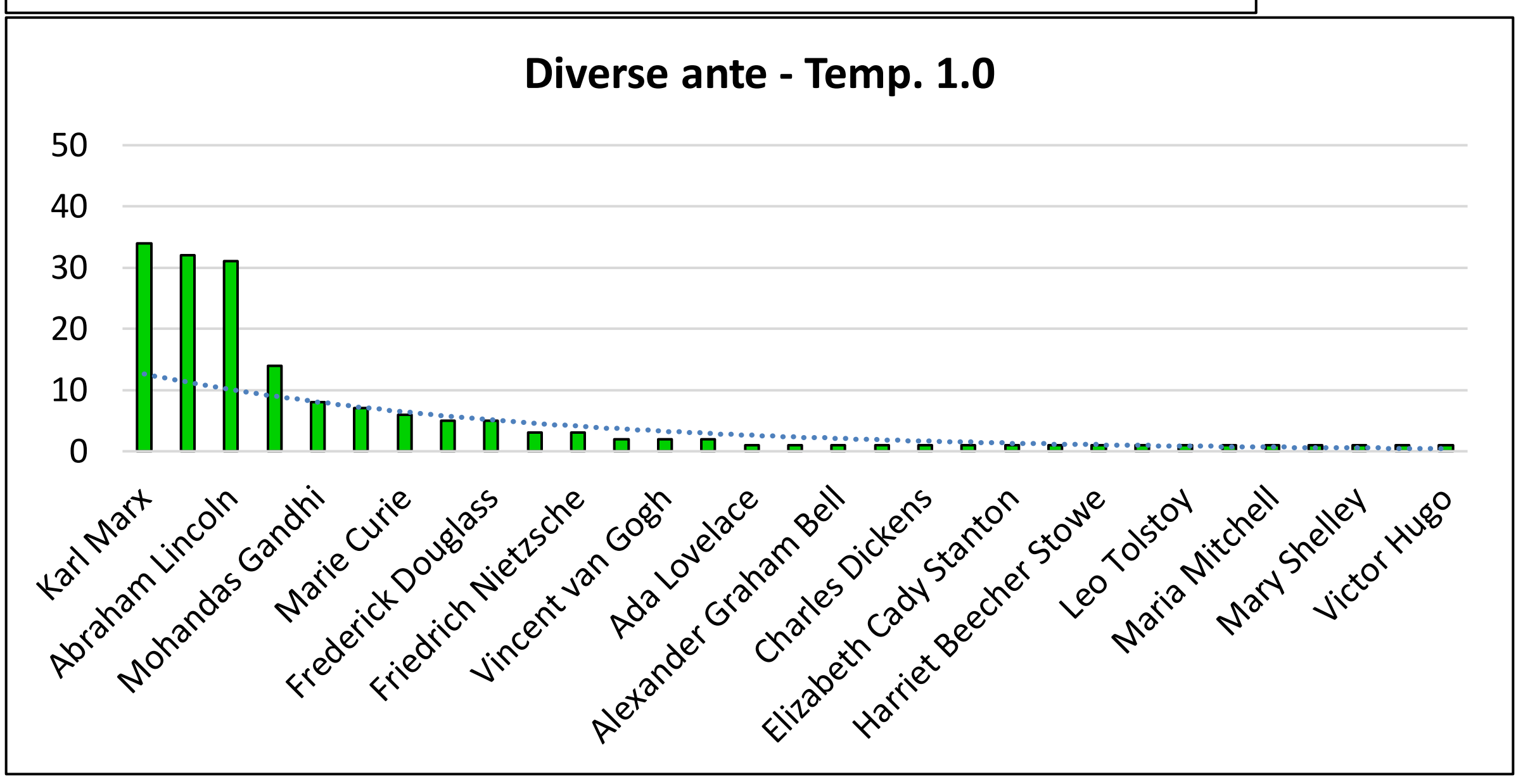

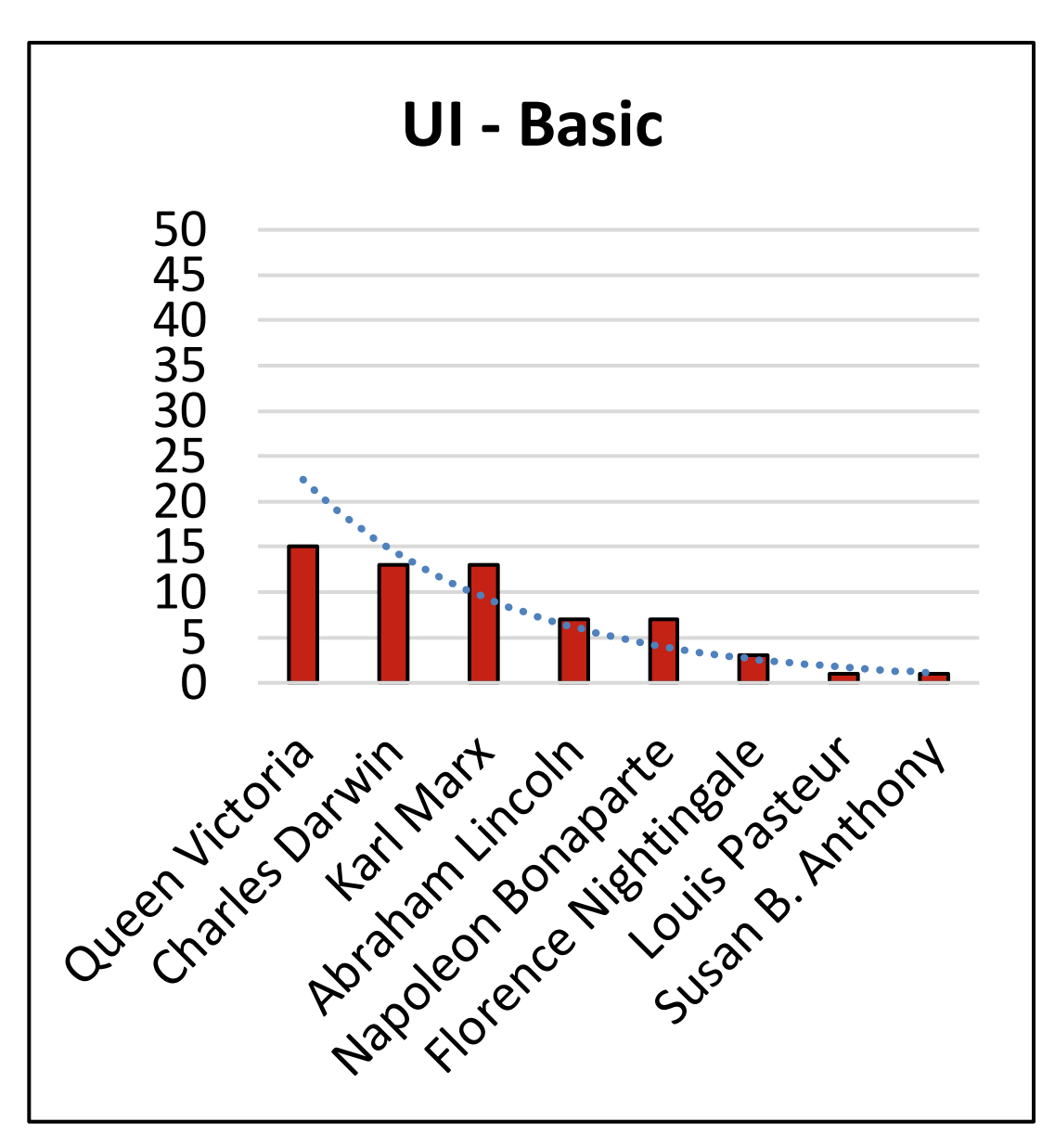

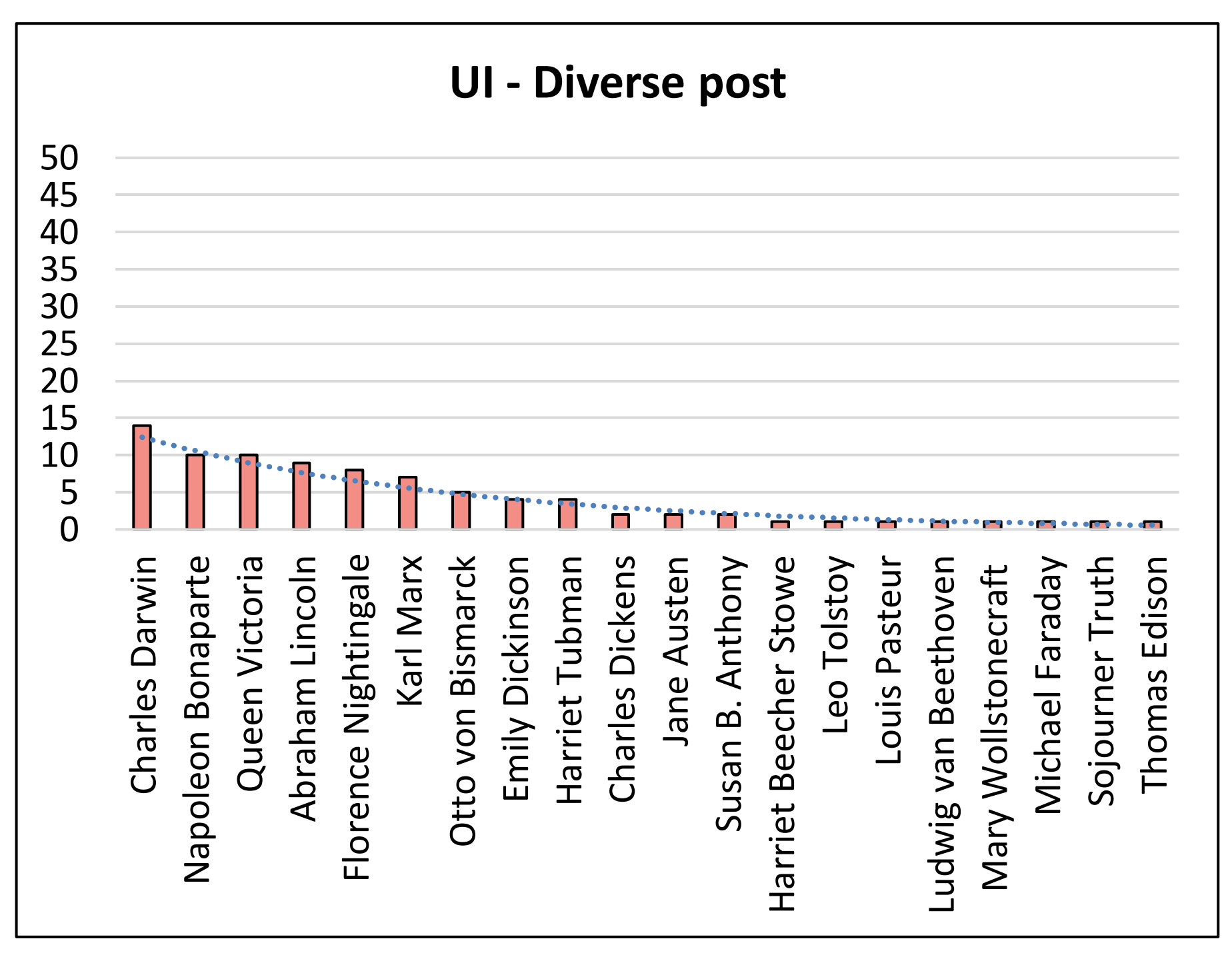

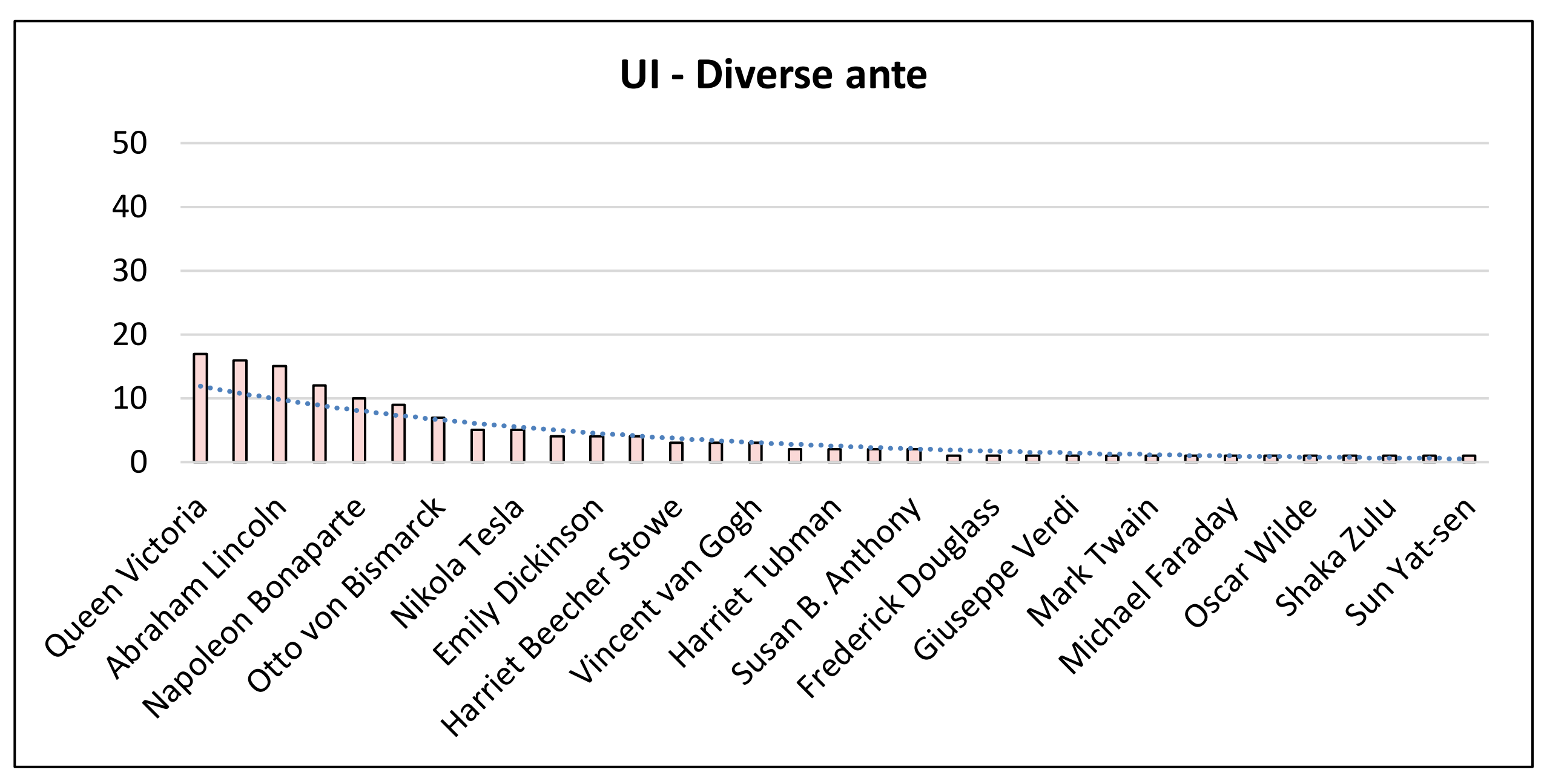

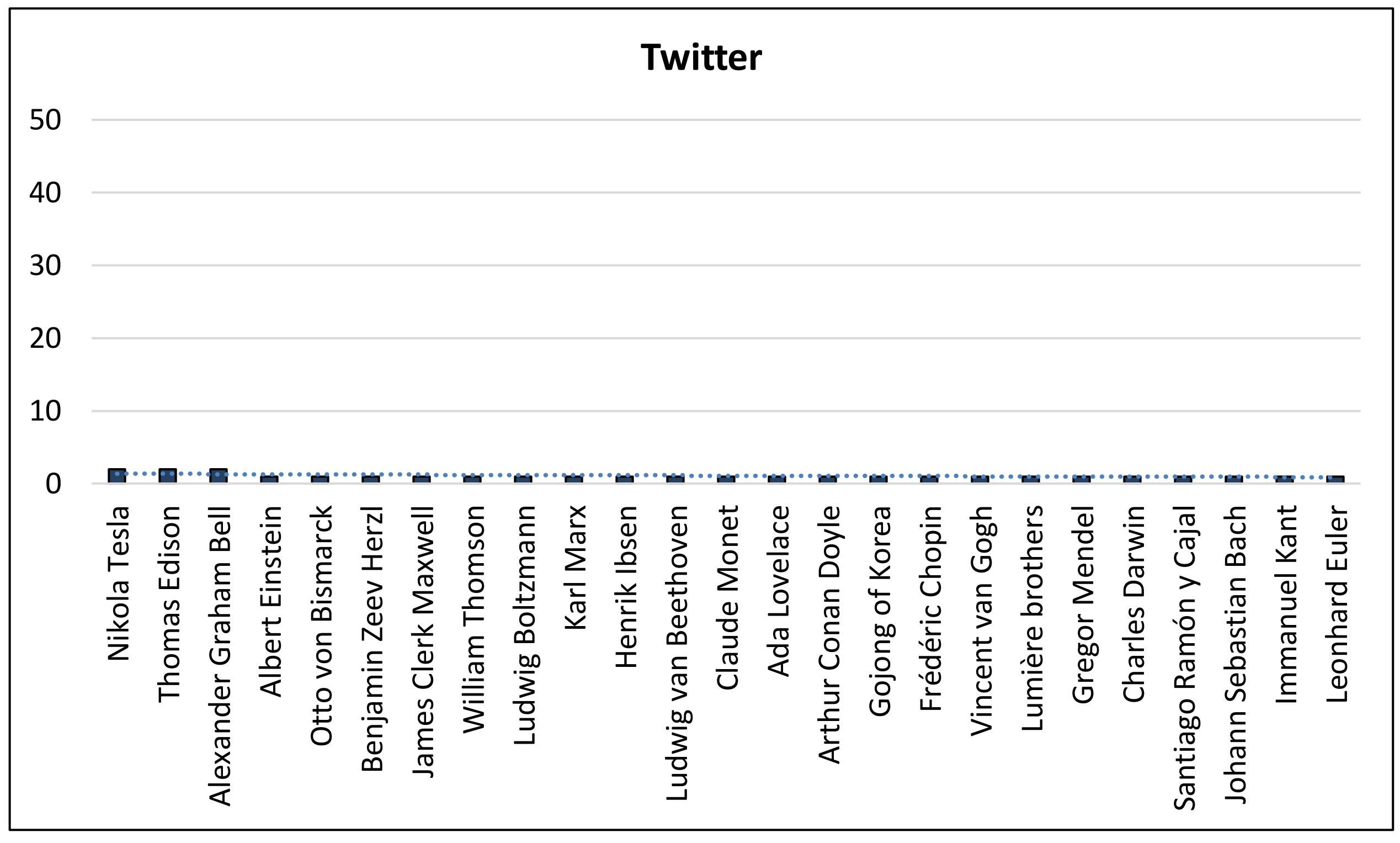

***Q2***

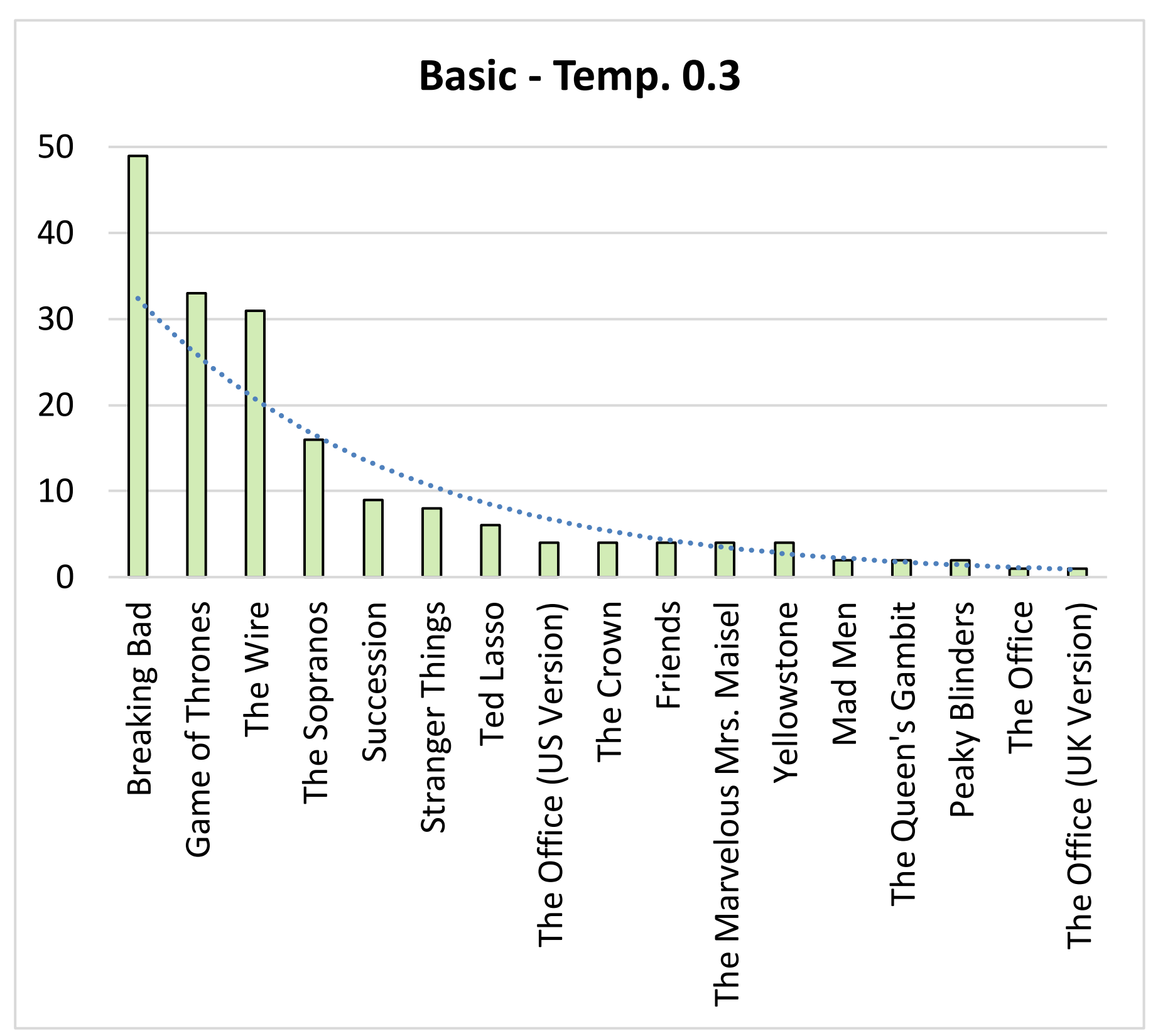

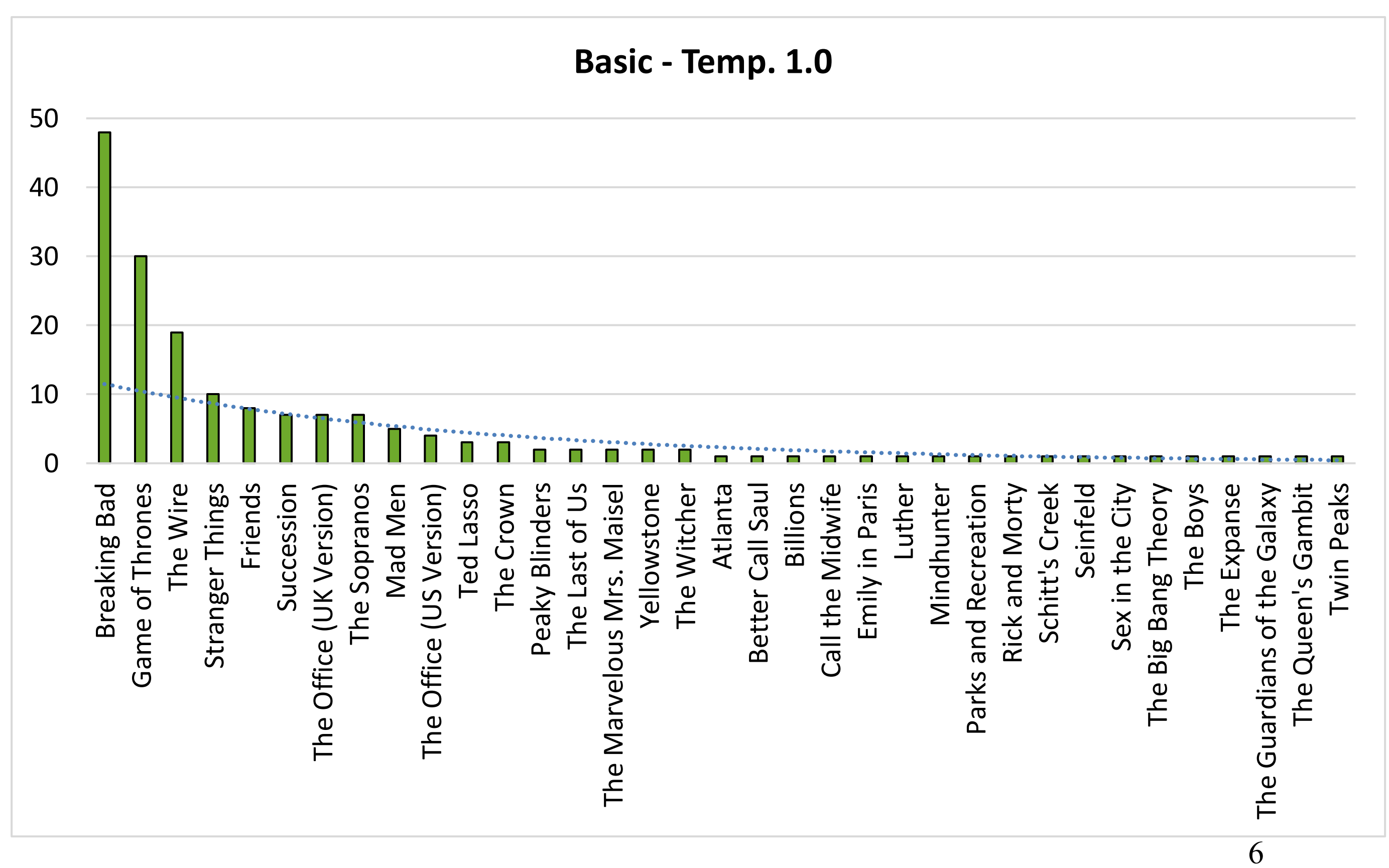

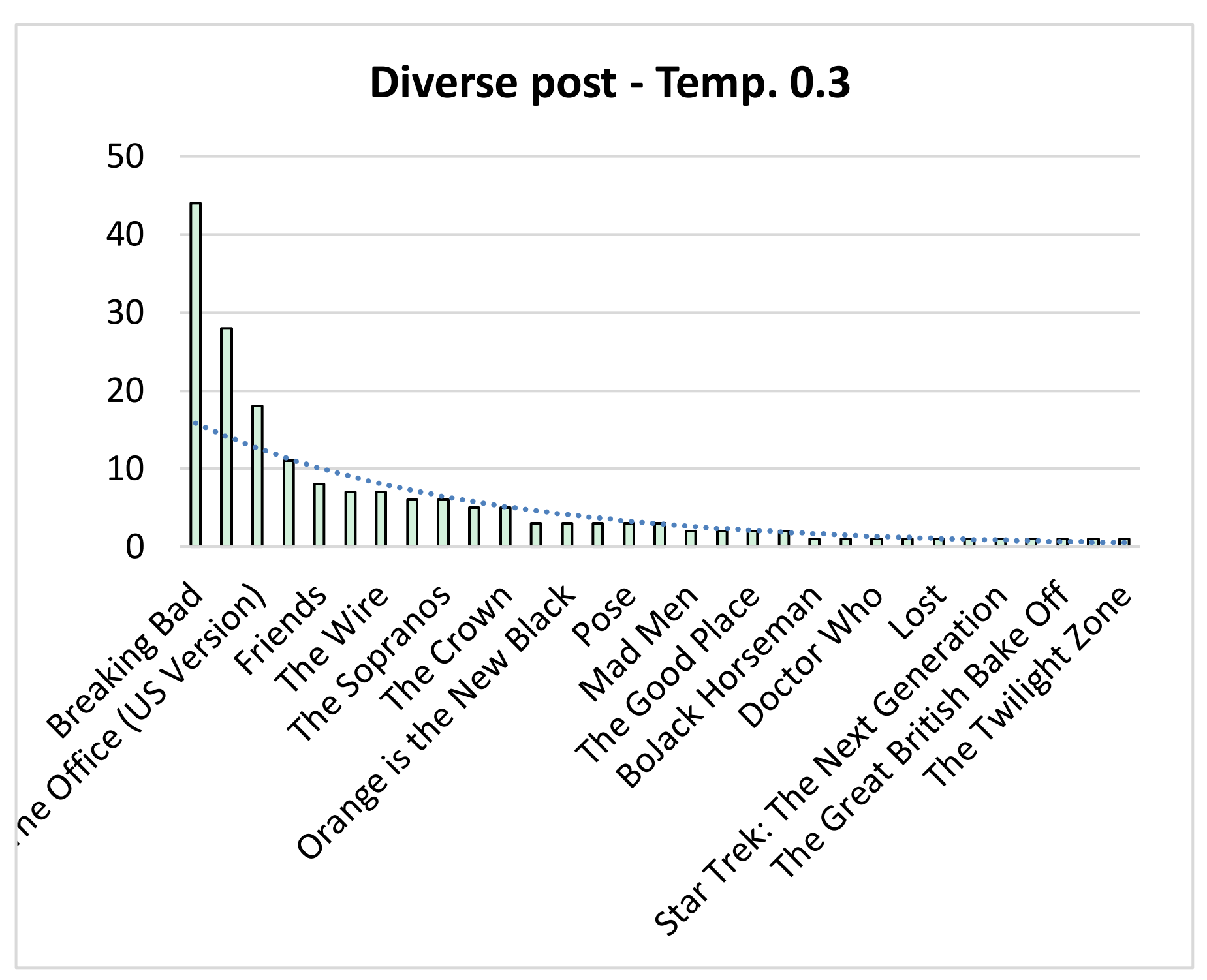

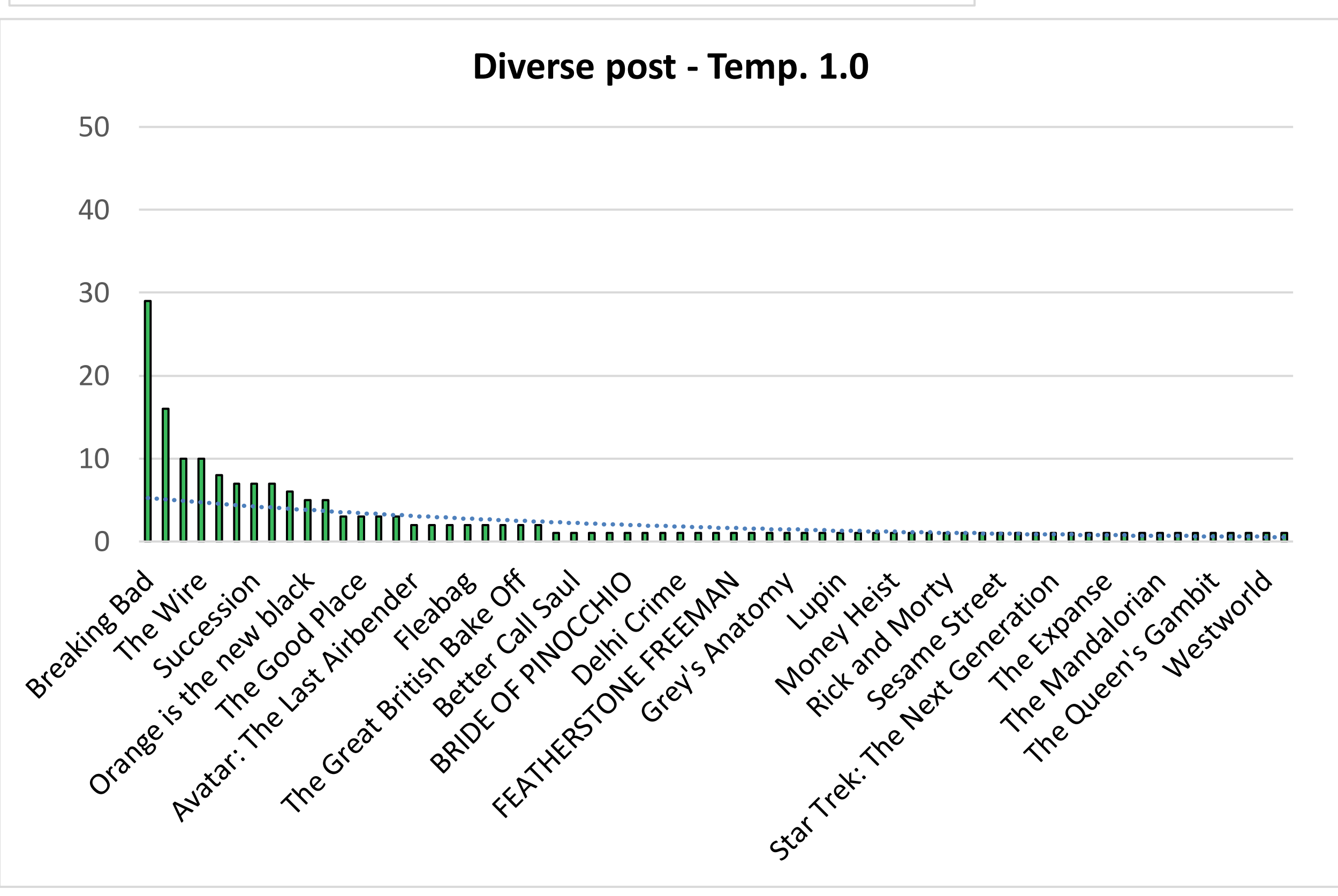

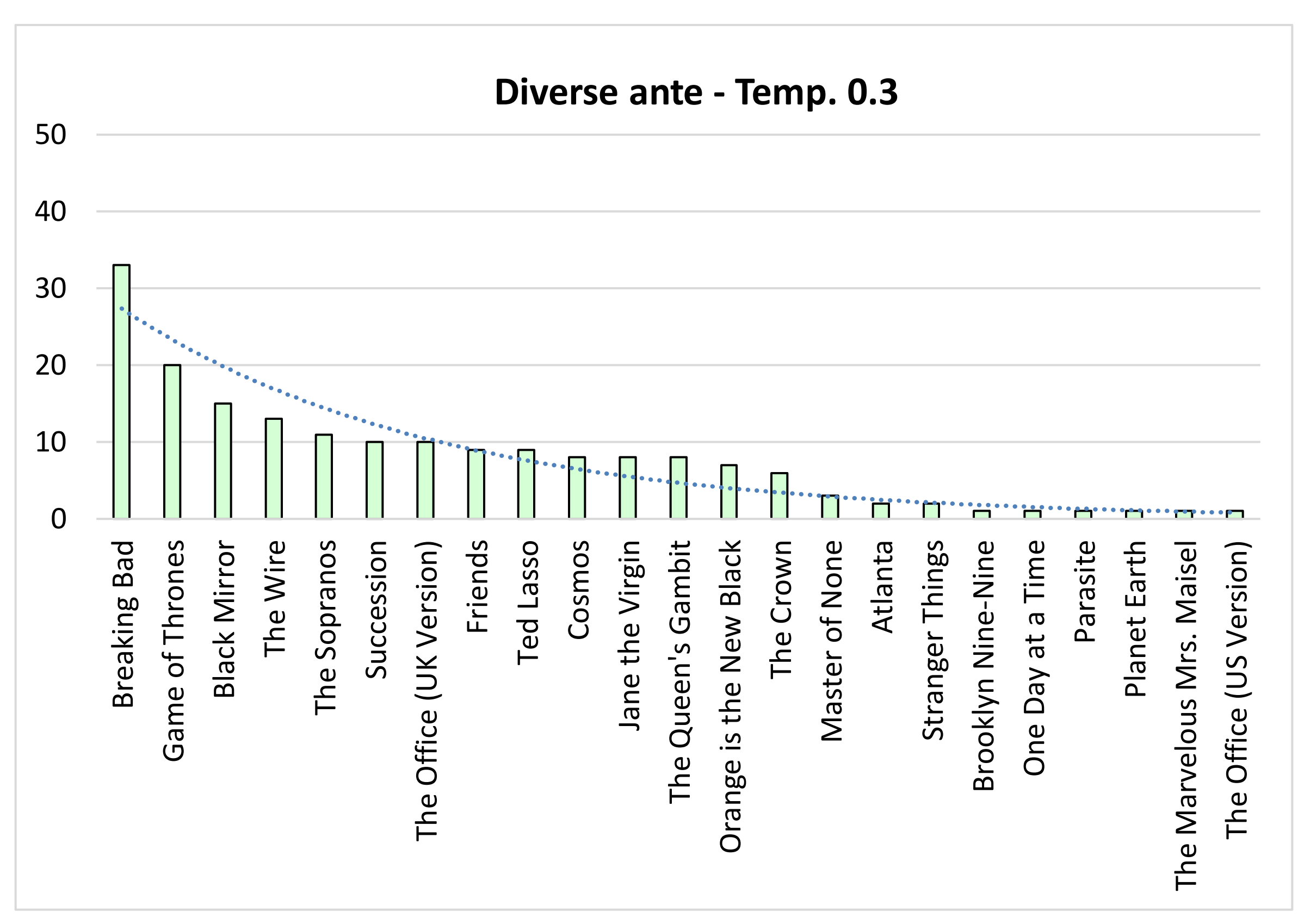

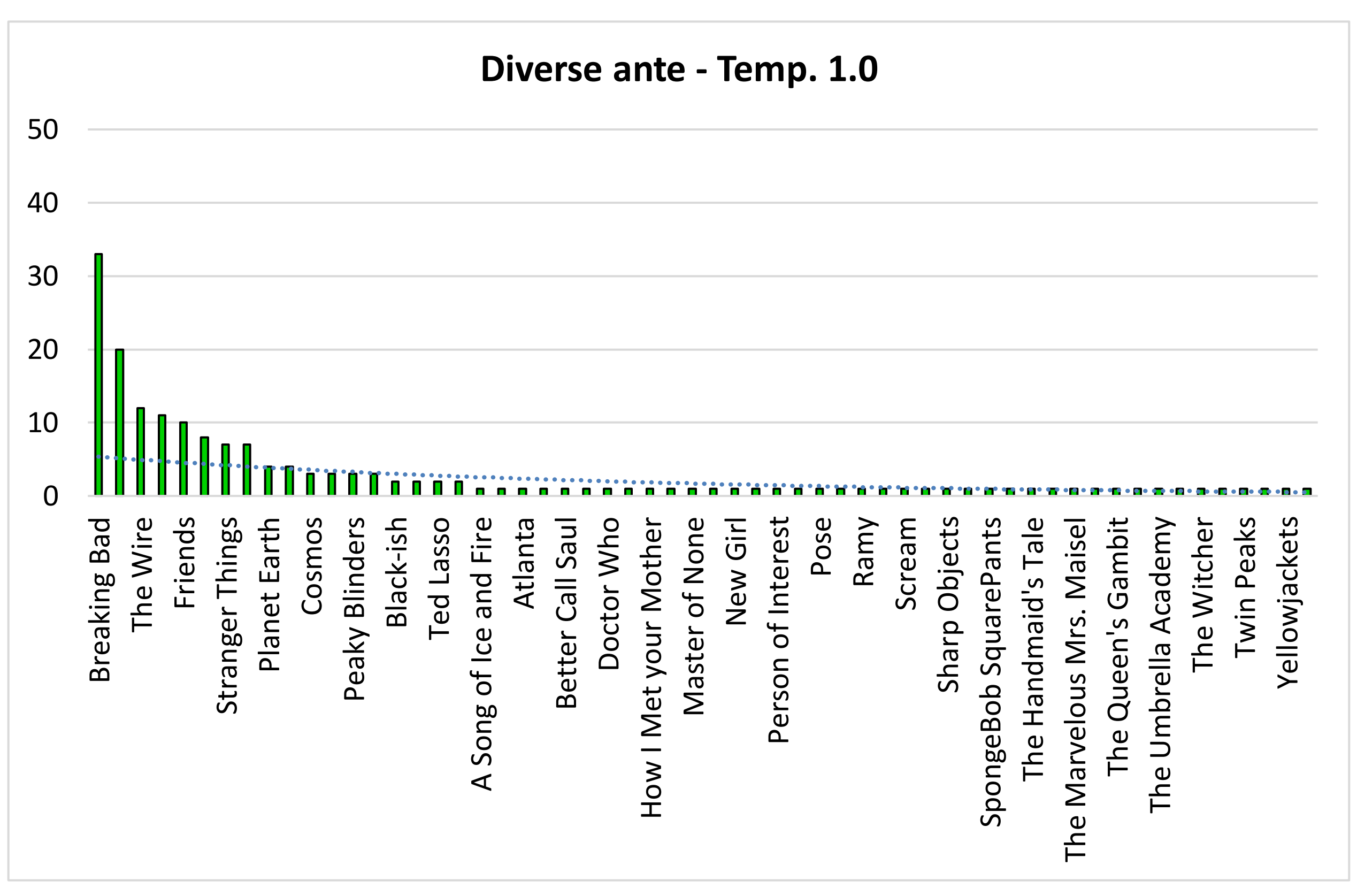

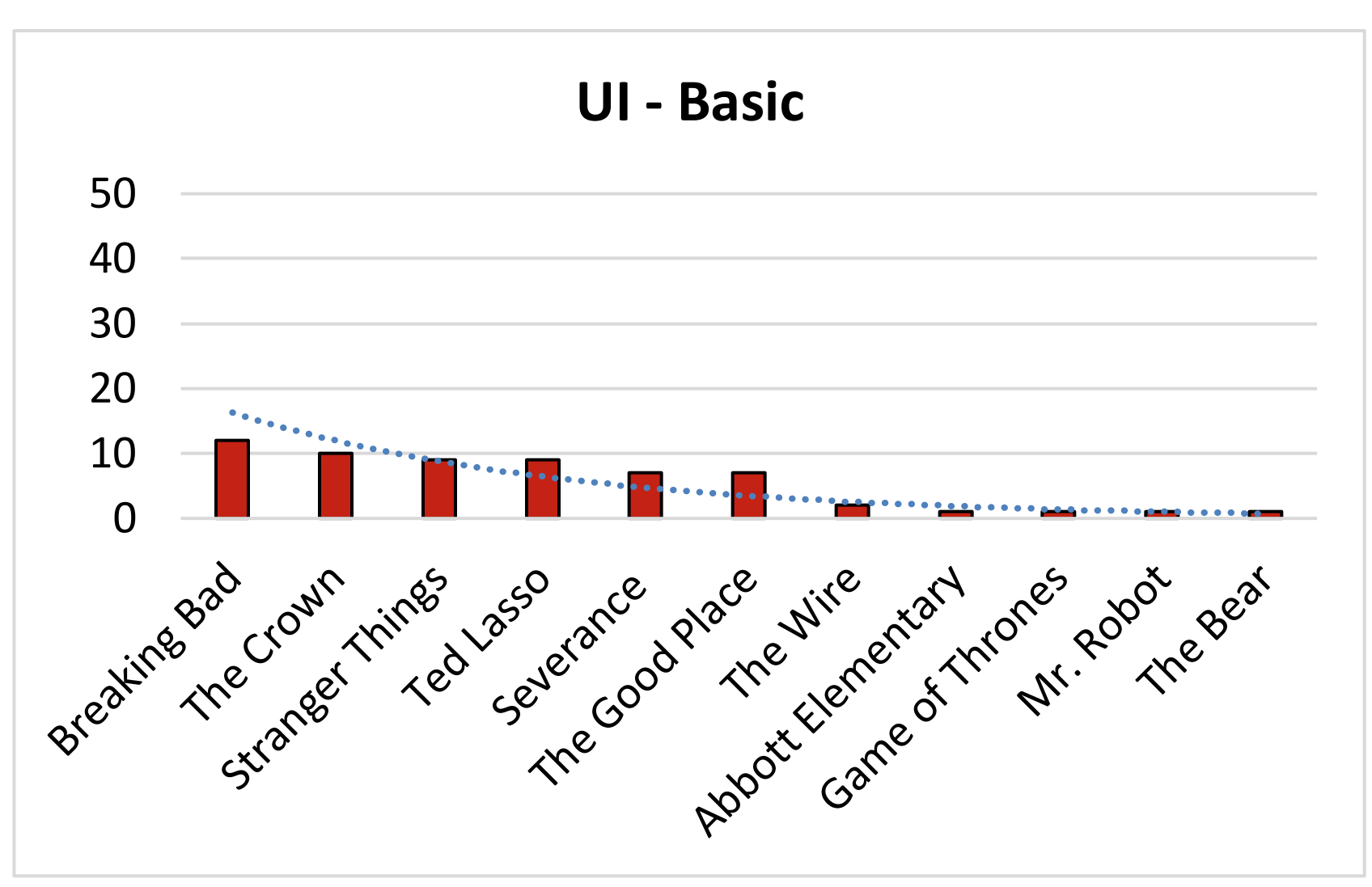

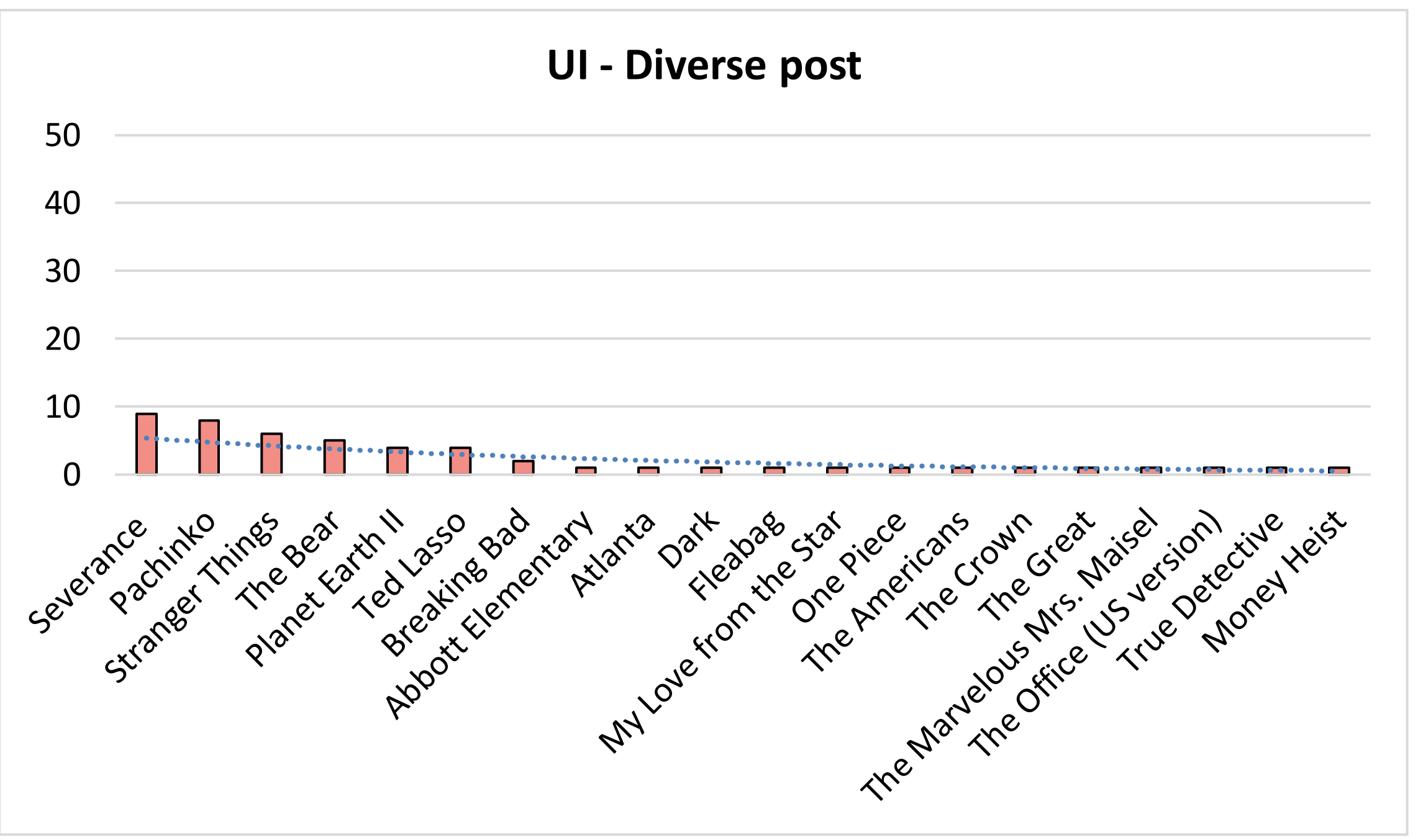

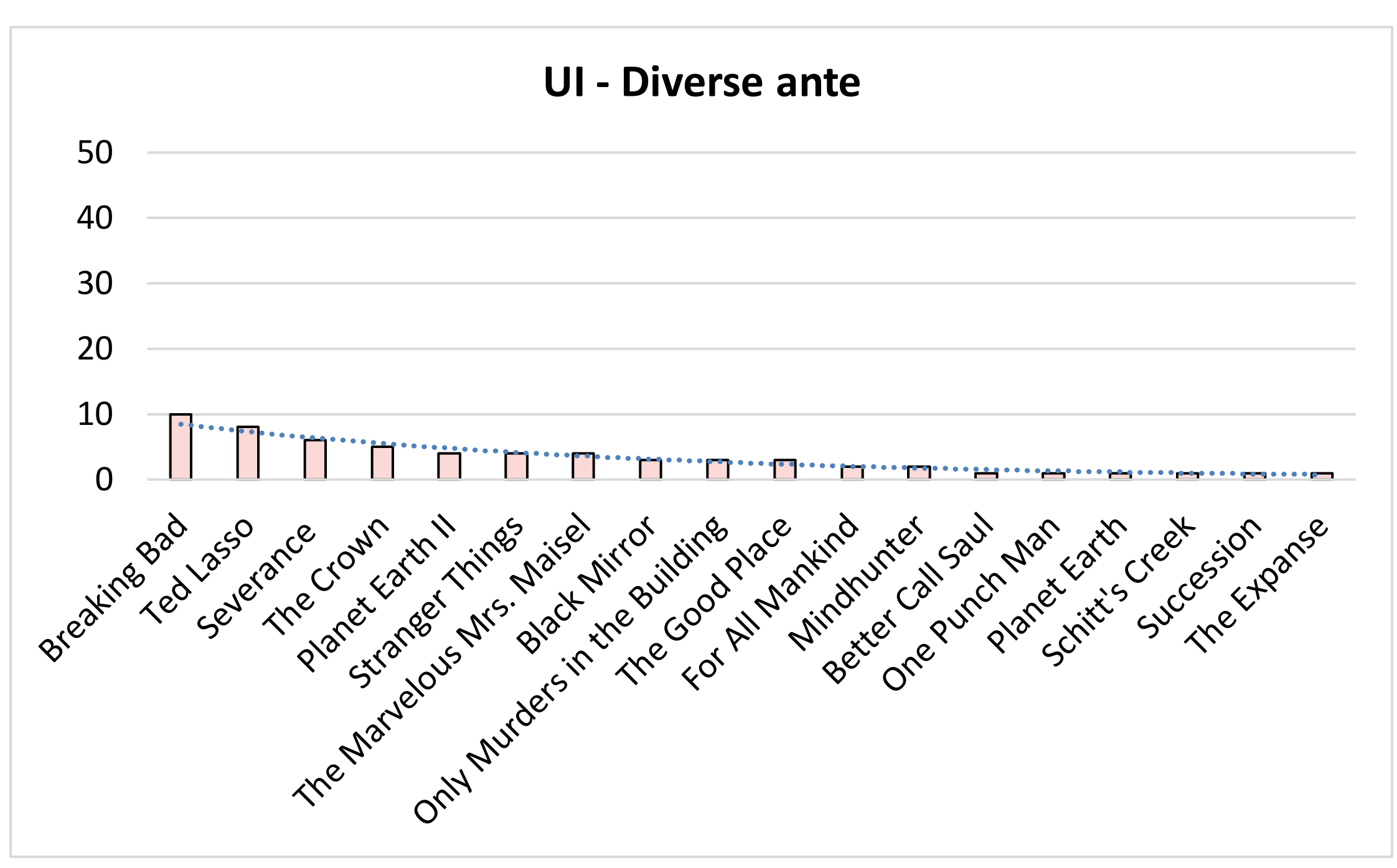

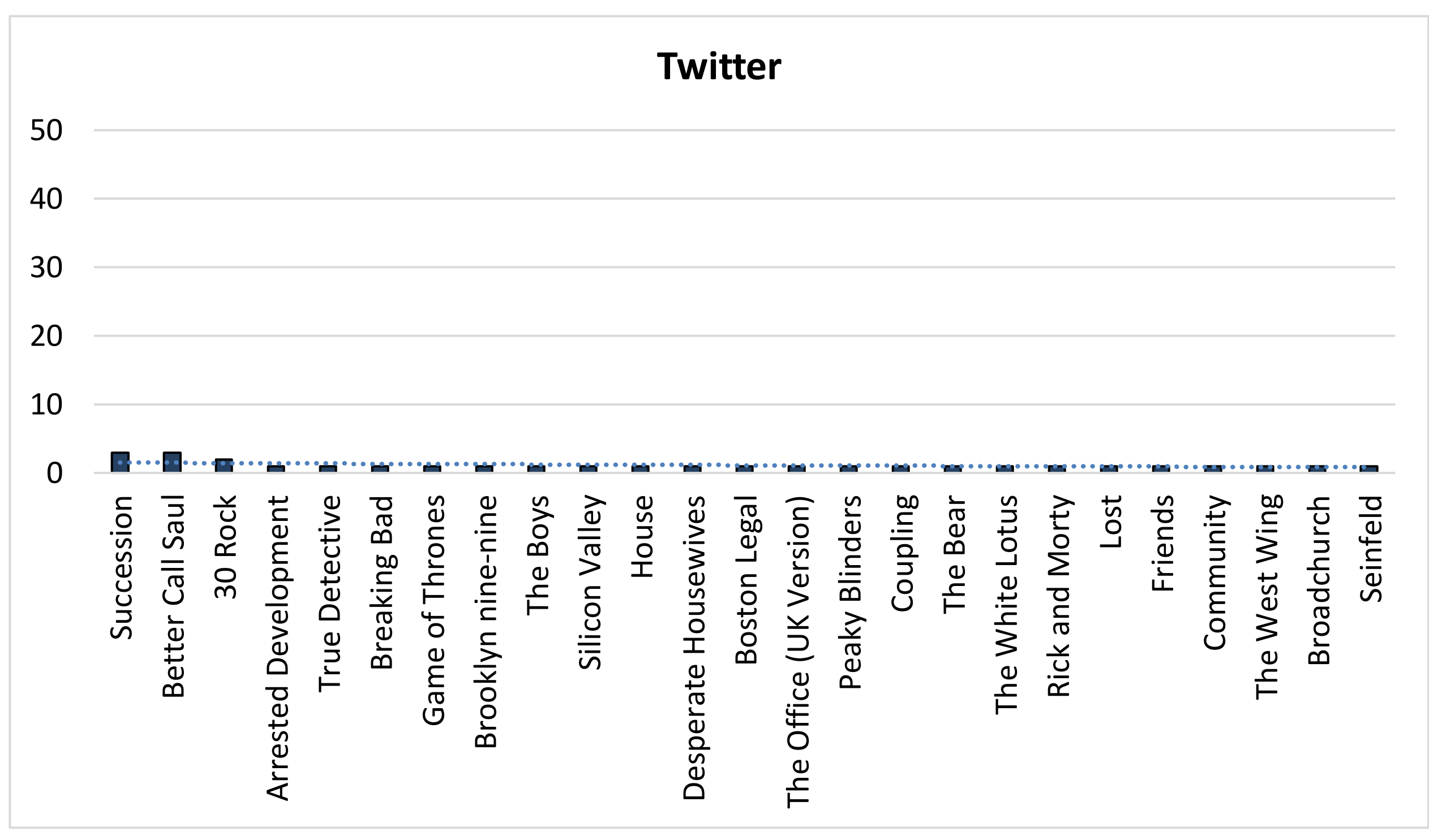

## *Q3*

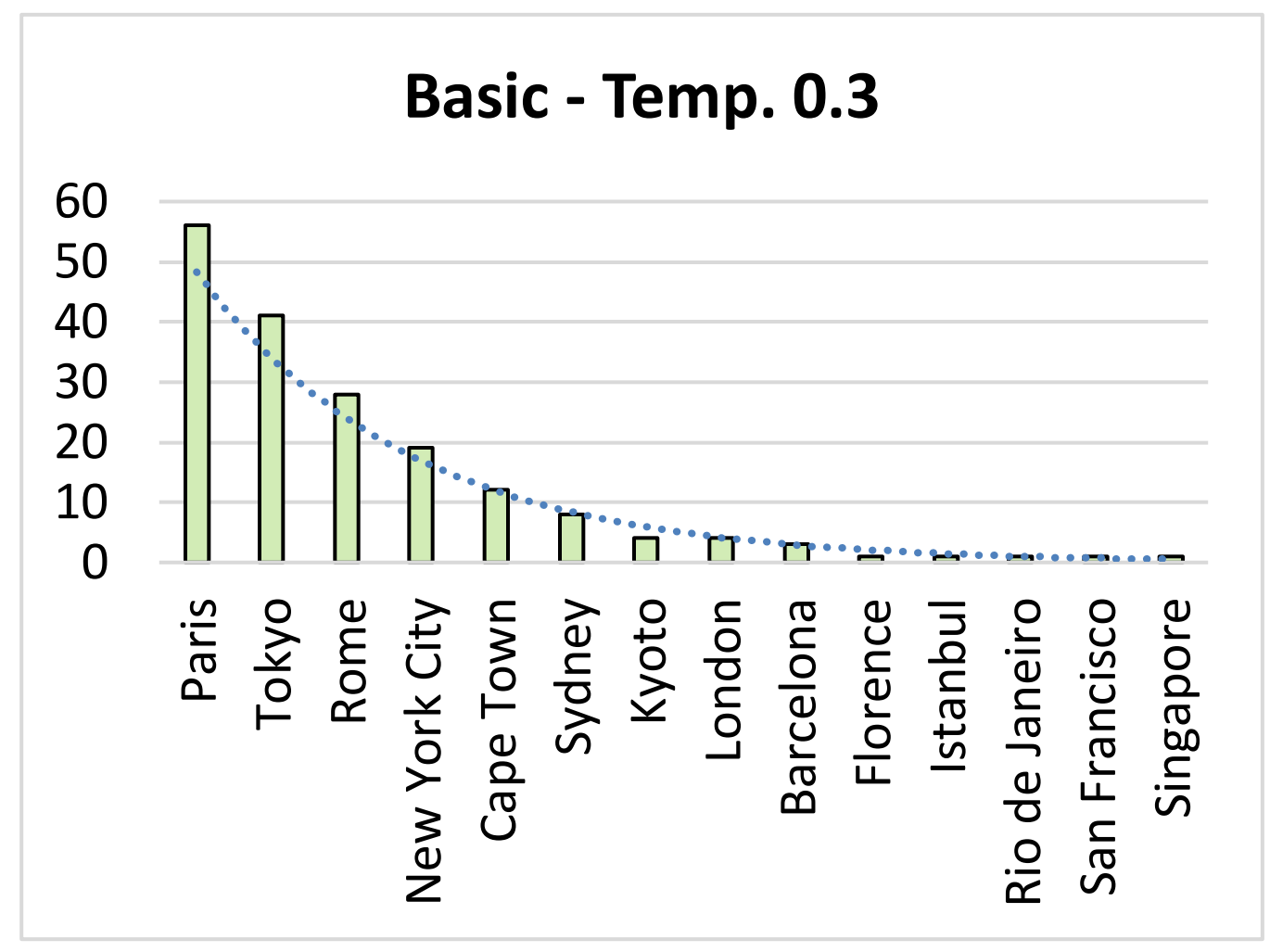

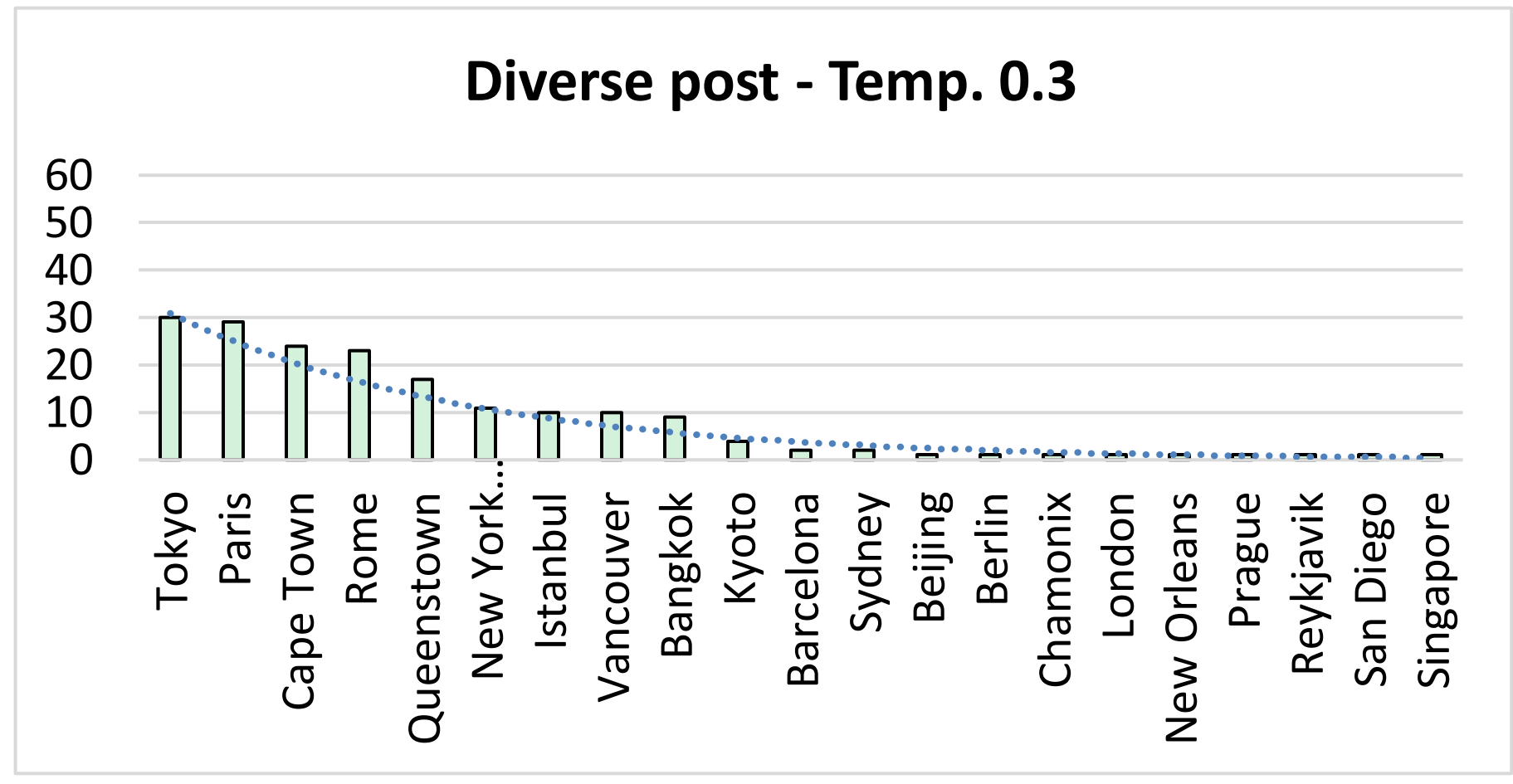

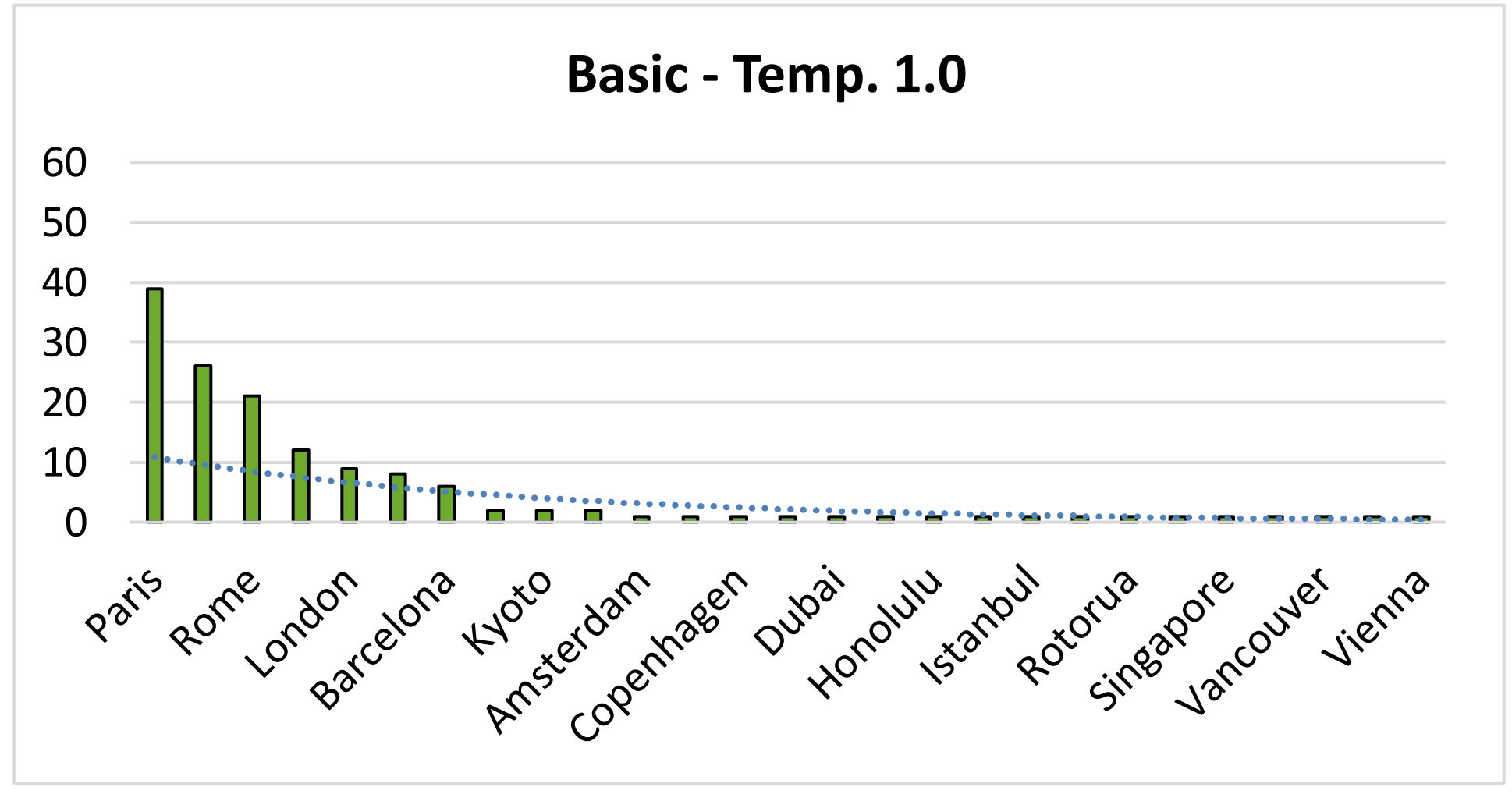

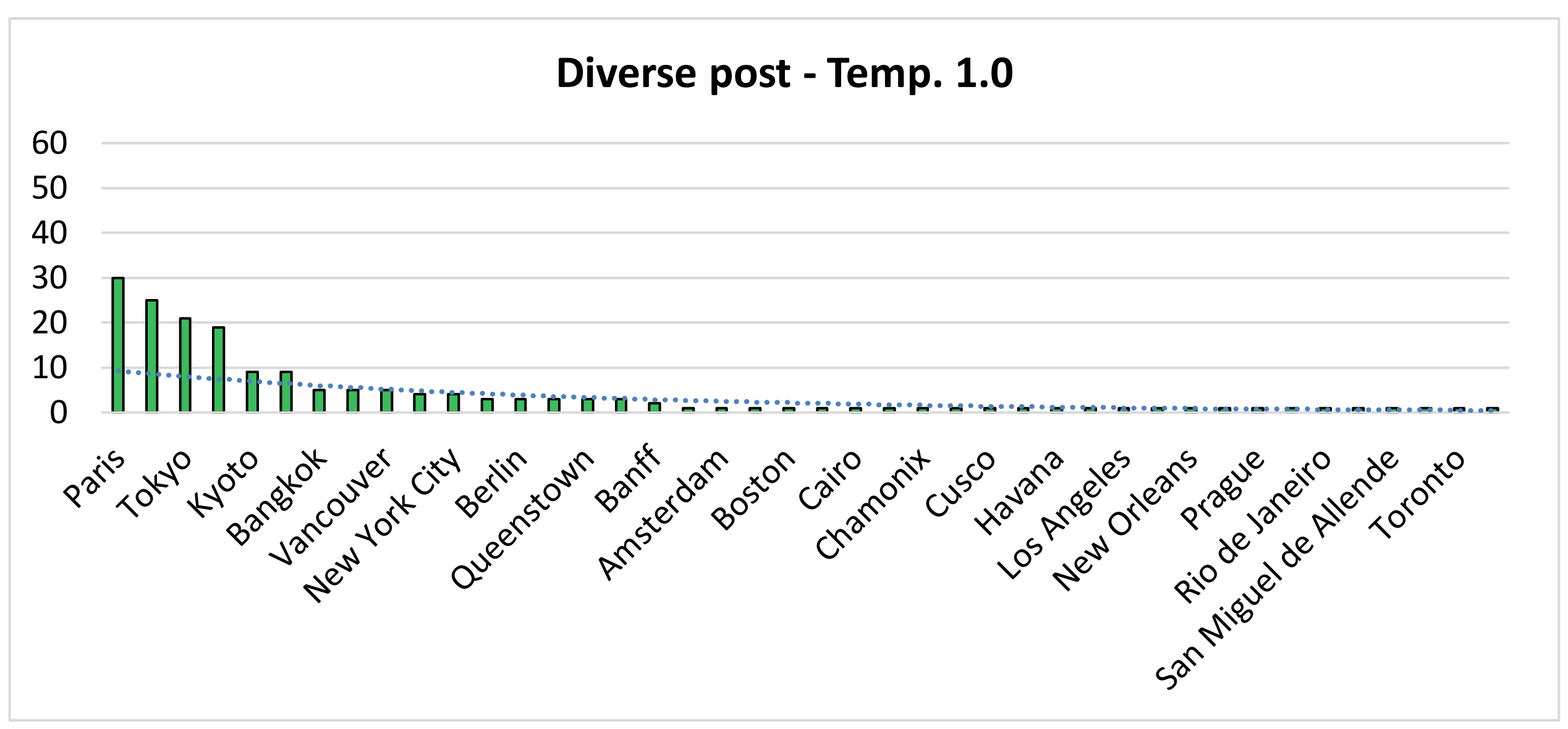

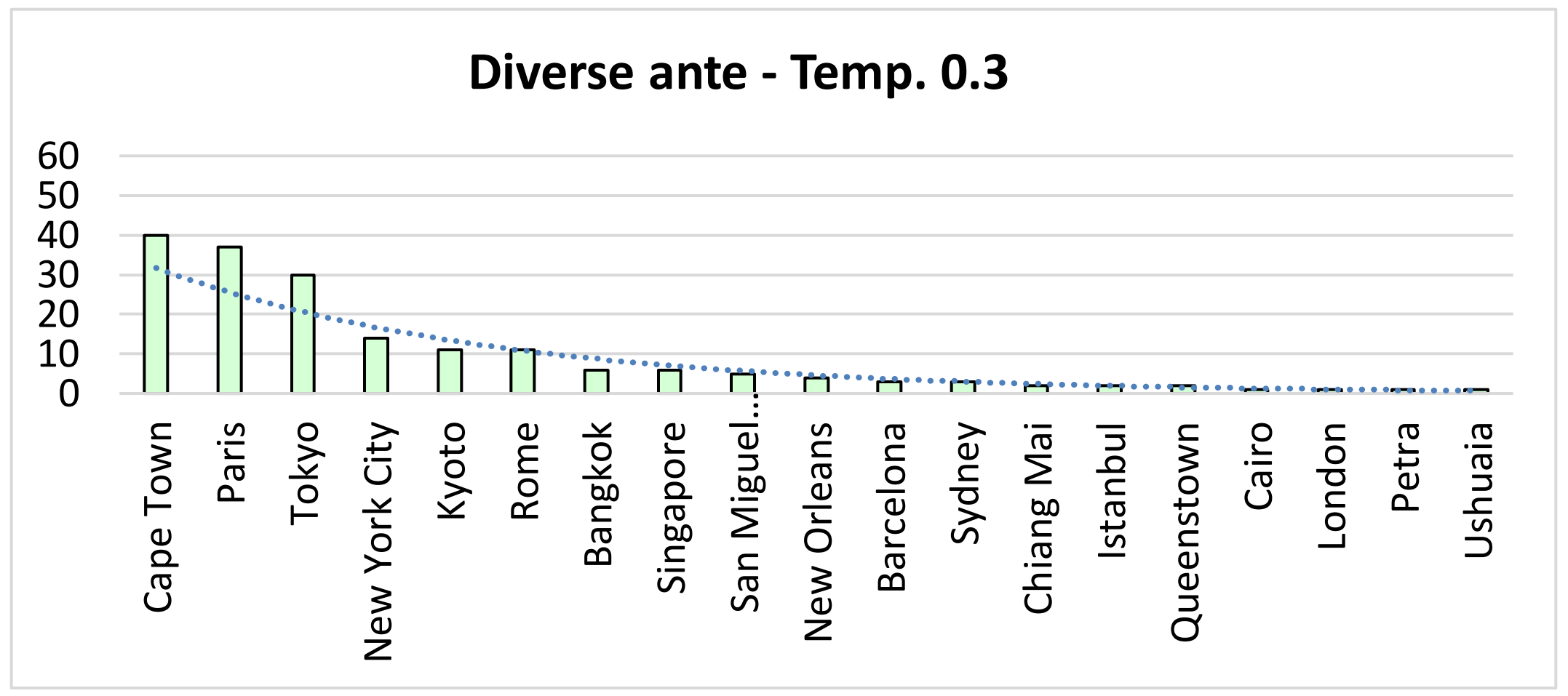

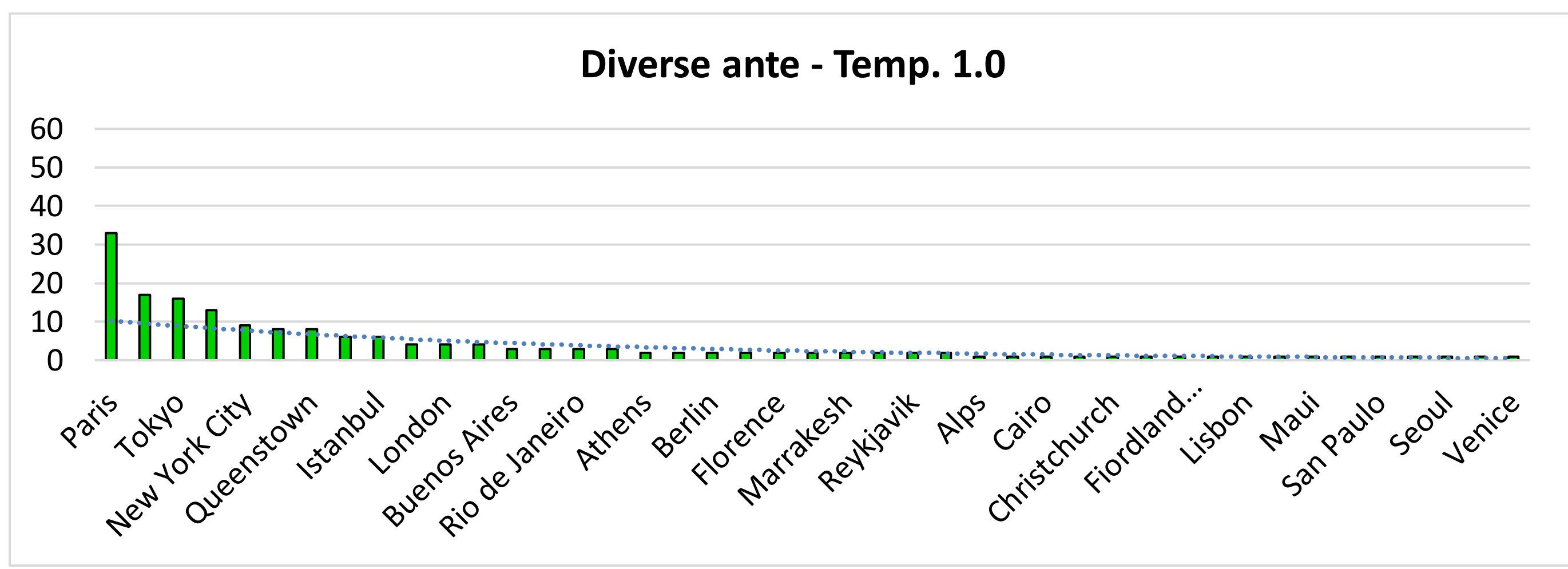

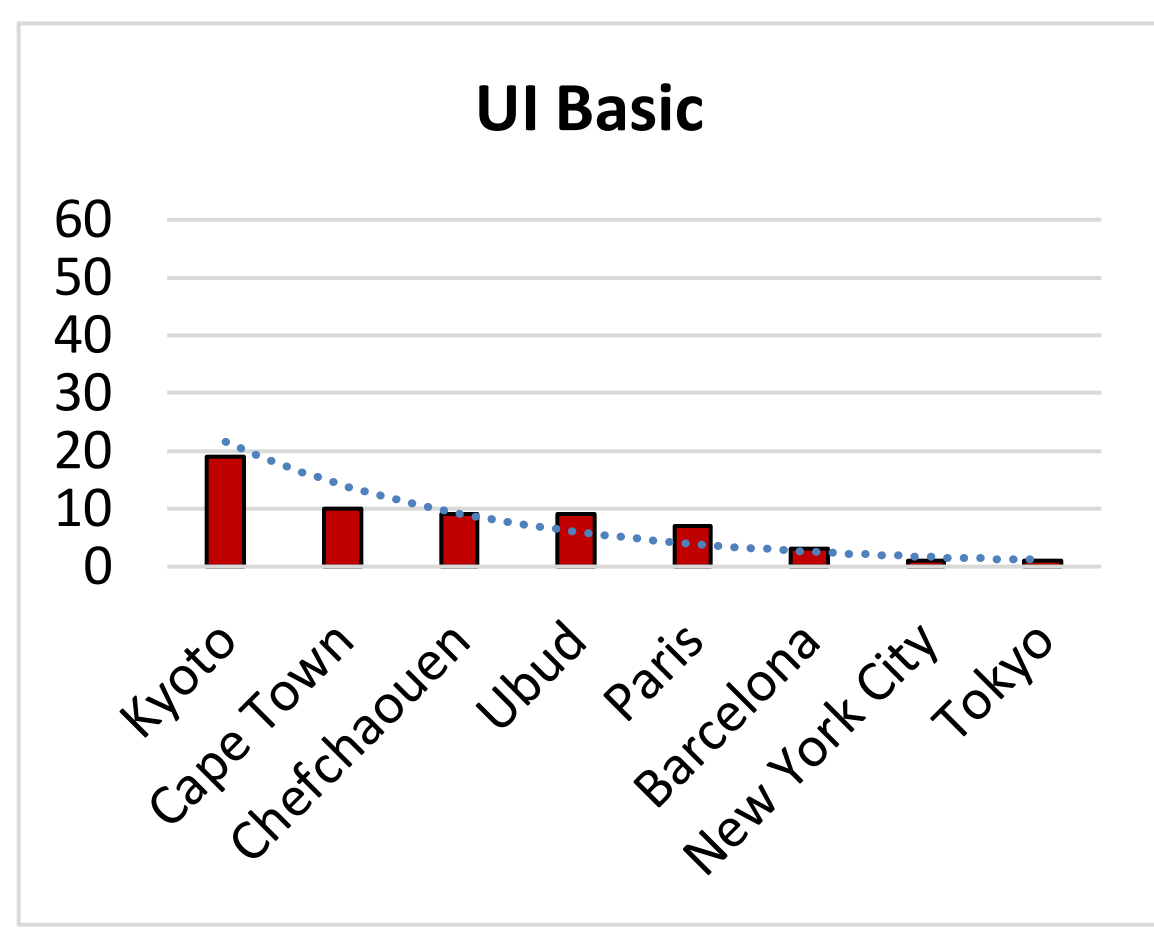

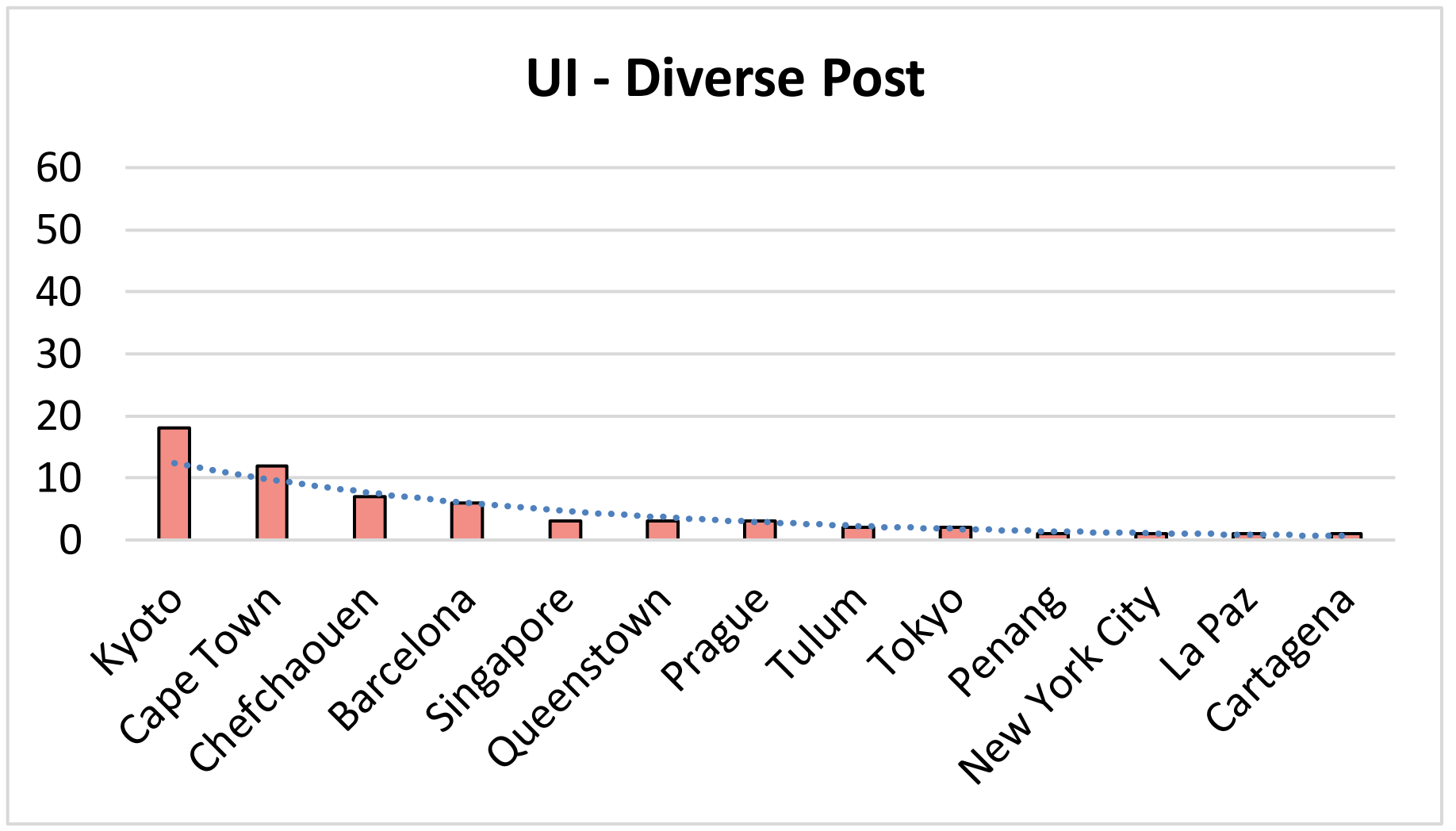

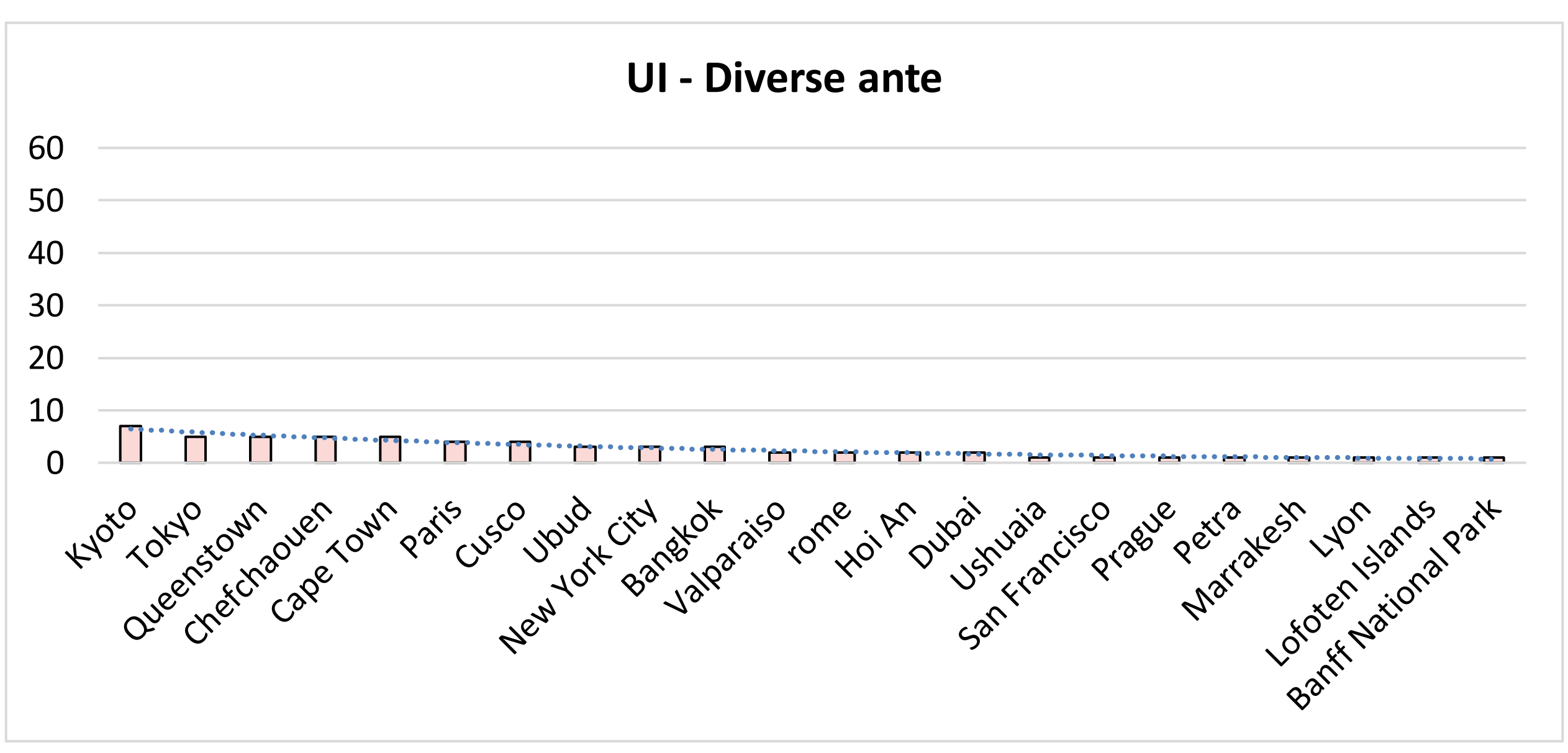

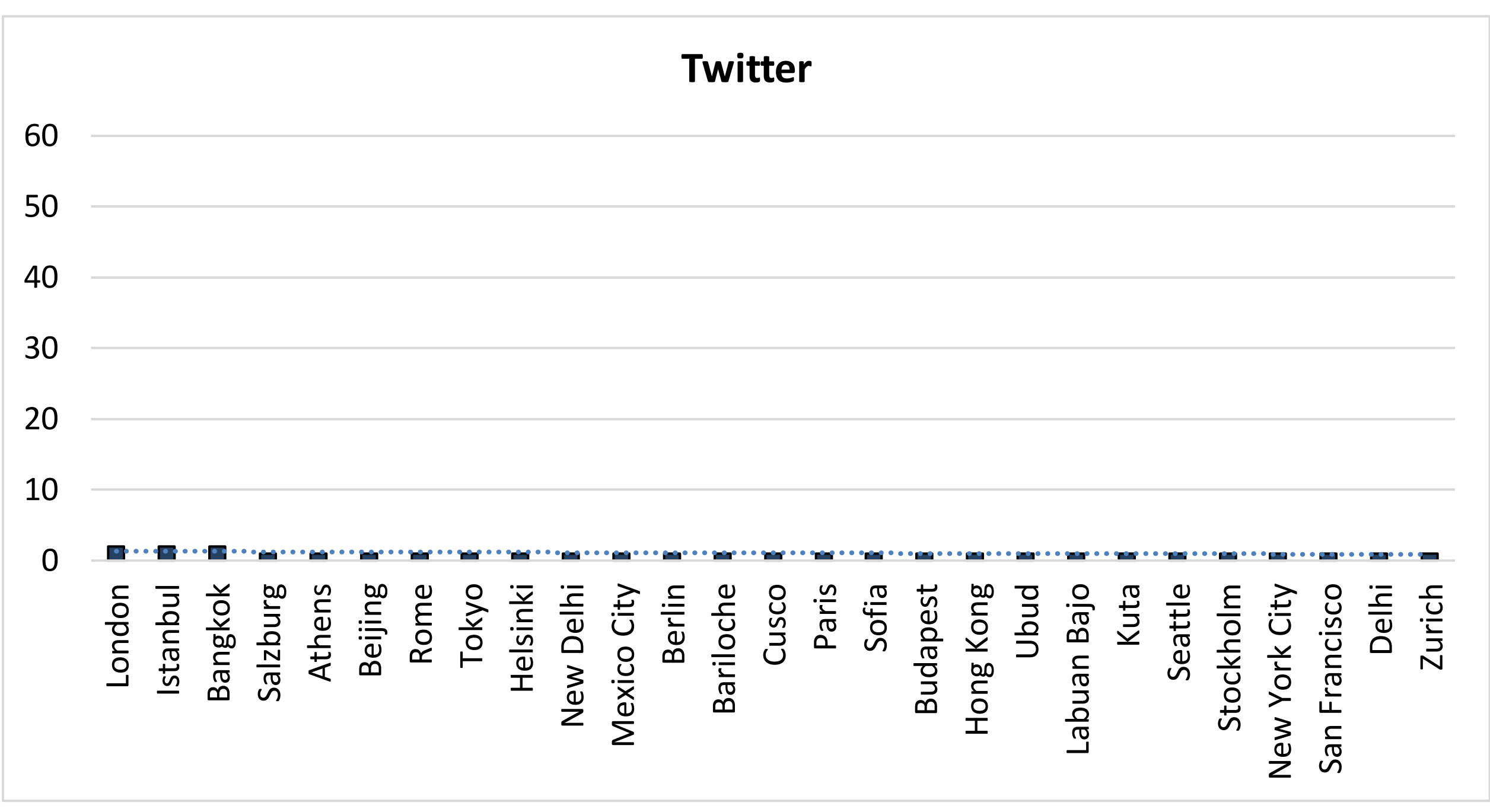

**Percentage of Errors in the Models' Outputs**

The table below presents the percentage of errors out of the total outputs of the models, under each of the different treatments.

| Q1 | |
|---|---|
| | Avg. Errors |
| Basic (with 0.3 Temperature) | **0%** |
| Basic (with 1.0 Temperature) | **2%** |
| Diverse post: (with 0.3 Temperature) | **0%** |
| Diverse post: (with 1.0 Temperature) | **1%** |
| Diverse Ante: (with 0.3 Temperature) | **0%** |
| Diverse Ante: (with 1.0 Temperature) | **1%** |
| Basic UI | **0%** |
| Diverse Post UI | **0%** |
| Diverse Ante UI | **0%** |

| Q2 | |
|---|---|
| | Avg. Errors |
| Basic (with 0.3 Temperature) | **0%** |
| Basic (with 1.0 Temperature) | **0%** |
| Diverse post: (with 0.3 Temperature) | **0%** |
| Diverse post: (with 1.0Temperature) | **1%** |
| Diverse Ante: (with 0.3 Temperature) | **1%** |
| Diverse Ante: (with 1.0 Temperature) | **1%** |
| Basic UI | **0%** |
| Diverse Post UI | **0%** |
| Diverse Ante UI | **0%** |

| Q3 | |
|---|---|
| | Avg. Errors |
| Basic (with 0.3 Temperature) | **0%** |
| Basic (with 1.0 Temperature) | **0%** |
| Diverse post: (with 0.3 Temperature) | **0%** |
| Diverse post: (with 1.0 Temperature) | **1%** |
| Diverse Ante: (with 0.3 Temperature) | **0%** |
| Diverse Ante: (with 1.0 Temperature) | **2%** |
| Basic UI | **0%** |
| Diverse Post UI | **0%** |
| Diverse Ante UI | **3%** |

# Part II – Raw Materials, Q1

https://github.com/technion-cs-nlp/growing-a-tail

# Part III - Raw Materials, Q2

https://github.com/technion-cs-nlp/growing-a-tail

# Part IV- Raw Materials, Q3

https://github.com/technion-cs-nlp/growing-a-tail